\documentclass[sigconf]{acmart}
\usepackage{spverbatim}
\usepackage{pifont}

\newcommand{\change}[1]{#1}

\AtBeginDocument{%
  \providecommand\BibTeX{{%
    \normalfont B\kern-0.5em{\scshape i\kern-0.25em b}\kern-0.8em\TeX}}}

\copyrightyear{2024}
\acmYear{2024}
\setcopyright{rightsretained}
\acmConference[UIST '24]{The 37th Annual ACM Symposium on User Interface Software and Technology}{October 13--16, 2024}{Pittsburgh, PA, USA}
\acmBooktitle{The 37th Annual ACM Symposium on User Interface Software and Technology (UIST '24), October 13--16, 2024, Pittsburgh, PA, USA}
\acmDOI{10.1145/3654777.3676401}
\acmISBN{979-8-4007-0628-8/24/10}
\begin{document}

\title{VoicePilot: Harnessing LLMs as Speech Interfaces for Physically Assistive Robots}


\author{Akhil Padmanabha\text{*}}
\affiliation{
  %\institution{Robotics Institute}
  \institution{Carnegie Mellon University}
  \city{Pittsburgh}
  \state{PA}
  \country{USA}
}
\email{akhilpad@andrew.cmu.edu}

\author{Jessie Yuan\text{*}}
\affiliation{
  %\institution{School of Computer Science}
  \institution{Carnegie Mellon University}
  \city{Pittsburgh}
  \state{PA}
  \country{USA}
}
\email{jzyuan@andrew.cmu.edu}

\author{Janavi Gupta}
\affiliation{
  %\institution{School of Computer Science}
  \institution{Carnegie Mellon University}
  \city{Pittsburgh}
  \state{PA}
  \country{USA}
}
\email{janavig@andrew.cmu.edu}

\author{Zulekha Karachiwalla}
\affiliation{
  %\institution{Robotics Institute}
  \institution{Carnegie Mellon University}
  \city{Pittsburgh}
  \state{PA}
  \country{USA}
}
\email{zkarachi@andrew.cmu.edu }

\author{Carmel Majidi}
\affiliation{
  %\institution{Department of Mechanical Engineering}
  \institution{Carnegie Mellon University}
  \city{Pittsburgh}
  \state{PA}
  \country{USA}
}
\email{cmajidi@andrew.cmu.edu }

\author{Henny Admoni}
\affiliation{
  %\institution{Robotics Institute}
  \institution{Carnegie Mellon University}
  \city{Pittsburgh}
  \state{PA}
  \country{USA}
}
\email{henny@cmu.edu}

\author{Zackory Erickson}
\affiliation{
  %\institution{Robotics Institute}
  \institution{Carnegie Mellon University}
  \city{Pittsburgh}
  \state{PA}
  \country{USA}
}
\email{zackory@cmu.edu}

\newcommand\blfootnote[1]{
    \begingroup
    \renewcommand\thefootnote{}\footnote{#1}
    \addtocounter{footnote}{-1}
    \endgroup
}

\renewcommand{\shortauthors}{Padmanabha*, Yuan* et al.}




\begin{abstract}
Physically assistive robots present an opportunity to significantly increase the well-being and independence of individuals with motor impairments or other forms of disability who are unable to complete activities of daily living. Speech interfaces, especially ones that utilize Large Language Models (LLMs), can enable individuals to effectively and naturally communicate high-level commands and nuanced preferences to robots. Frameworks for integrating LLMs as interfaces to robots for high level task planning and code generation have been proposed, but fail to incorporate human-centric considerations which are essential while developing assistive interfaces. In this work, we present a framework for incorporating LLMs as speech interfaces for physically assistive robots, constructed iteratively with 3 stages of testing involving a feeding robot, culminating in an evaluation with 11 older adults at an independent living facility. We use both quantitative and qualitative data from the final study to validate our framework and additionally provide design guidelines for using LLMs as speech interfaces for assistive robots. Videos, code, and supporting files are located on our project website\footnote{\url{https://sites.google.com/andrew.cmu.edu/voicepilot/}}
\end{abstract}


%\keywords{Do, Not, Us, This, Code, Put, the, Correct, Terms, for,Your, Paper}

%  \includegraphics[width=\textwidth]{figures/teaser.pdf}
%  \vspace{-0.8cm}
%  \caption{Top left: Our work introduces an iteratively constructed framework for integrating LLMs as speech interfaces for assistive robots. Bottom: We develop a speech interface using the framework and evaluate it with 11 older adult participants at an independent living facility. Top right: Using insights from the study, we present a reflection and 5 design guidelines.}
%  \Description{TODO}

\begin{teaserfigure}
\vspace{-0.5cm}
 \includegraphics[width=\textwidth]{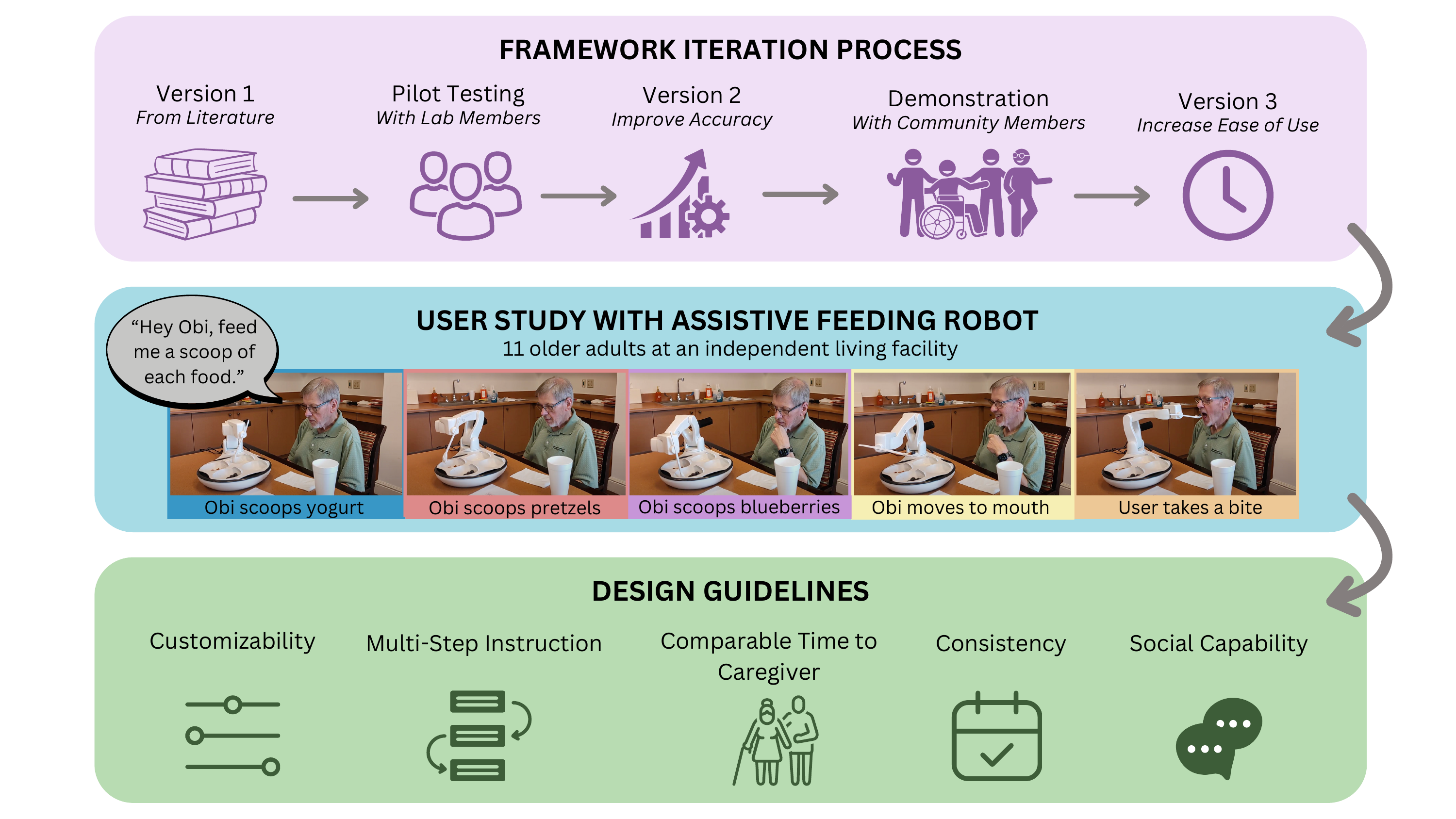}
  \vspace{-0.8cm}
  \caption{Top: Our work introduces \change{a framework that was iteratively developed} for integrating LLMs as speech interfaces for assistive robots. Middle: We develop a speech interface for a feeding robot using the framework and evaluate it with 11 older adults at an independent living facility. Bottom: Using insights from the study, we present a reflection and 5 design guidelines.}
  \Description{The top section of the figure shows icons that represent the different phases of the iteratively constructed framework. For Version 1 from literature, the icon is a set of books. For Pilot Testing with lab members, the icon is 3 people. For Version 2 Improve Accuracy, the icon is a graph with an arrow that is sloped upwards. For Demonstration with community members, the icon is a set of 4 people, one of whom is in a wheelchair. For Version 3 Increase Ease of Use, the icon is a clock. In the middle section of the figure, labelled User Study with Assistive Feeding Robot, 11 older adults at an independent living facility, an older adult says "Hey Obi, feed me a scoop of each food" and there are a series of 5 pictures that show Obi scooping the yogurt, then scooping the pretzels, then scooping the blueberries, then moving to mouth, and lastly the user taking a bite. Lastly, on the bottom section of the figure, titled Design Guidelines, 5 icons are shown for the design guidelines: customizability, multi-step instruction, comparable time to caregiver, consistency, and social capability.}
  \label{fig:teaser}
\end{teaserfigure}


\maketitle

\section{Introduction}
Motor impairments affect a significant percentage of the United States with approximately 5 million individuals (1.7\%) affected by varying degrees of paralysis due to conditions including stroke and spinal cord injury~\cite{armour2016prevalence}. Physical impairments and other forms of disabilities can hinder individuals from performing activities of daily living (ADLs), such as eating and bathing, and instrumental activities of daily living (iADLs), such as cooking and cleaning. This can significantly impact an individual's independence and quality of life, requiring them to rely on a caregiver for assistance~\cite{qol1, qol2, qo13, qol4}. This is especially the case in older populations with over 20\% of individuals in the United States over 65 years of age requiring assistance with at least one of their self care or mobility tasks~\cite{freedman2014behavioral}. 

\begin{figure}[t!]
  \centering
  \includegraphics[width = \columnwidth]{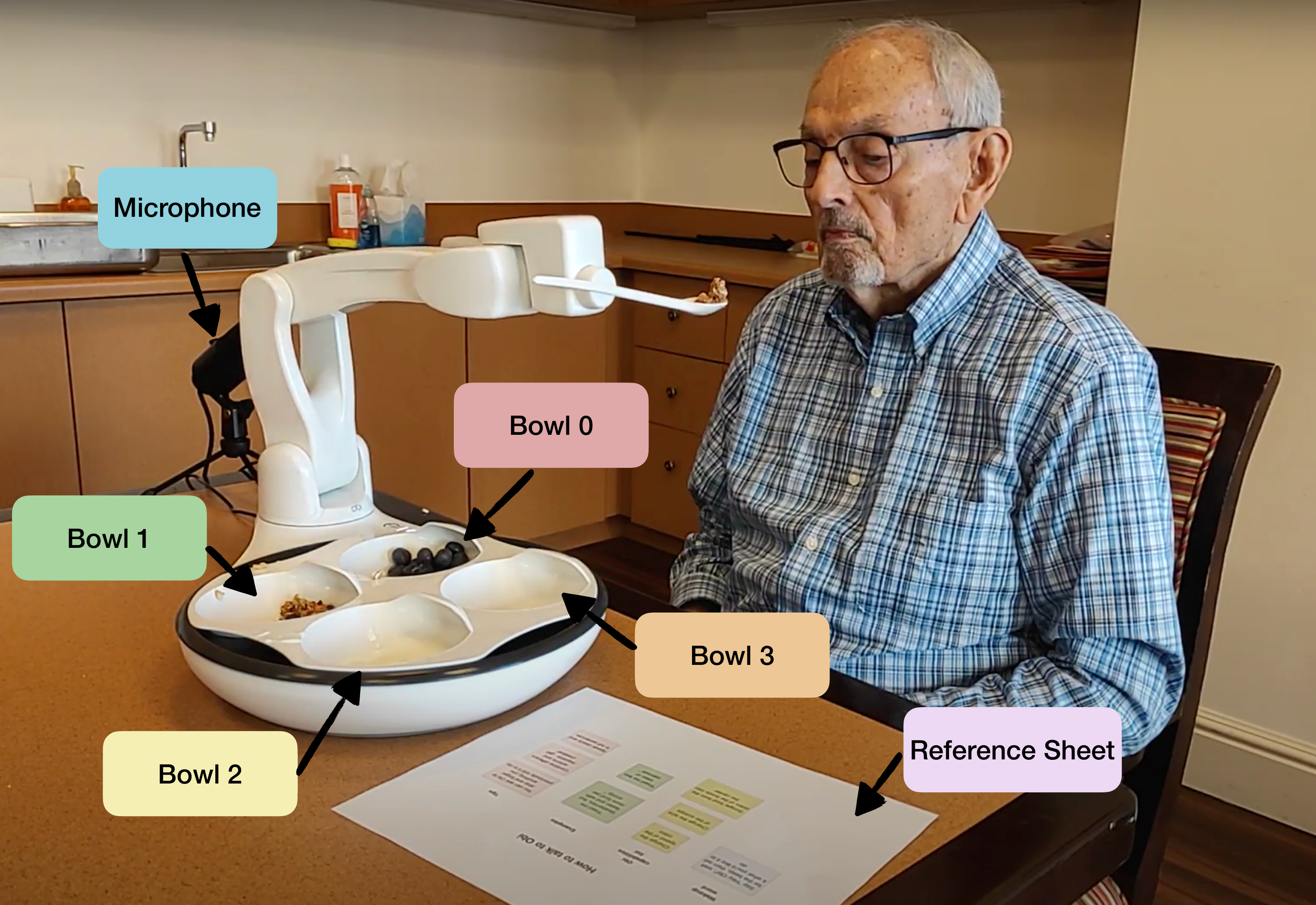}
  \caption{Obi robot and setup for study with 11 older adults at an independent living facility. The Obi robot's arm moves towards a participant's mouth with a spoonful of granola. A microphone is positioned to the right of the participant and a cheat sheet with example commands is placed to the left of the participant.}
  \Description{An older adult from our human study at an independent living facility is shown with the Obi robot bringing a spoonful of granola towards his mouth. A reference sheet and microphone are placed next to the Obi robot. The bowls of the robot are labelled. Bowl 0 contains blueberries. Bowl 1 contains granola. Bowl 2 contains yogurt. Bowl 3 is empty.}
  \label{fig:obi_study3_setup}
\end{figure}

Physically assistive robots, such as the adaptive eating robot Obi~\cite{Obi}, shown in Fig.~\ref{fig:obi_study3_setup}, can enable individuals with impairments to perform a range of self-care and household tasks~\cite{tasks, tasksaroundhead, king2012dusty, nanavati2023physically, yang2023high}. Assistive interfaces can allow individuals with varying forms of disability to control such robots. Interfaces often have physical requirements, dictating which individuals can effectively use them. For example, individuals without the ability to move their head may use eye tracking with a web-based interface~\cite{robotsforhumanity, TapoMaya} while those with head motion may prefer inertial interfaces that capture their head movements and directly map them to robot motion~\cite{padmanabha2023hat, padmanabha2024independence}. 

Among these interfaces, speech excels as an option for those with the ability to speak, as it allows individuals to naturally provide robots with both high-level commands and nuanced preferences or customizations. Speech interfaces for assistive robots have been explored extensively for control of both mobile robots and robot arms~\cite{pulikottil2018voice, lauretti2017comparative, house2009voicebot} and can be more intuitive than existing interface options like joysticks~\cite{poirier2019voice}. With the advent of Large Language Models (LLMs) and ongoing research combining LLMs with robots, robust speech interfaces for robots are becoming increasingly viable in the near future. Existing works have presented frameworks and guidelines for integrating LLMs with robots for high-level task planning and code generation~\cite{singh2023progprompt, liang2023code, vemprala2023chatgpt, brohan2023rt, zhang2023large, zhao2023chat, ahn2024autort, wang2024lami, mahadevan2024generative}. \change{These works concentrate on prompt engineering and code generation to enhance the accuracy of large language models (LLMs). No prior works involve human subjects interacting directly with an LLM-integrated robot for physically assistive tasks. Consequently, they lack insight and guidelines on the human-centric considerations essential for using LLMs as assistive interfaces.}

To address this gap, we develop a framework and design guidelines for integrating LLMs as speech interfaces for physically assistive robots. Our presented framework underwent extensive iterative development \change{and consists of components related to prompt engineering and system rollout, the process of deploying the LLM-based speech interface and robot with users.} First, we leveraged insight from existing literature to develop an initial version. We applied this Version 1 framework to integrate an existing LLM, GPT-3.5 Turbo from OpenAI~\cite{gptturbo}, with the commercial feeding robot Obi~\cite{Obi}. After testing this interface with lab members, we refined the framework, iterated on our LLM-based speech interface for the Obi, and demonstrated it to community members at a disability awareness event (Version 2). We used insights from the demonstration to develop a final framework and interface, evaluated through a user study involving 11 older adults at an independent living facility (Version 3). Lastly, we use both quantitative and qualitative data from the study with older adults to present design guidelines. 

\begin{figure*}[hbt!]
  \centering
  \includegraphics[width = \textwidth]{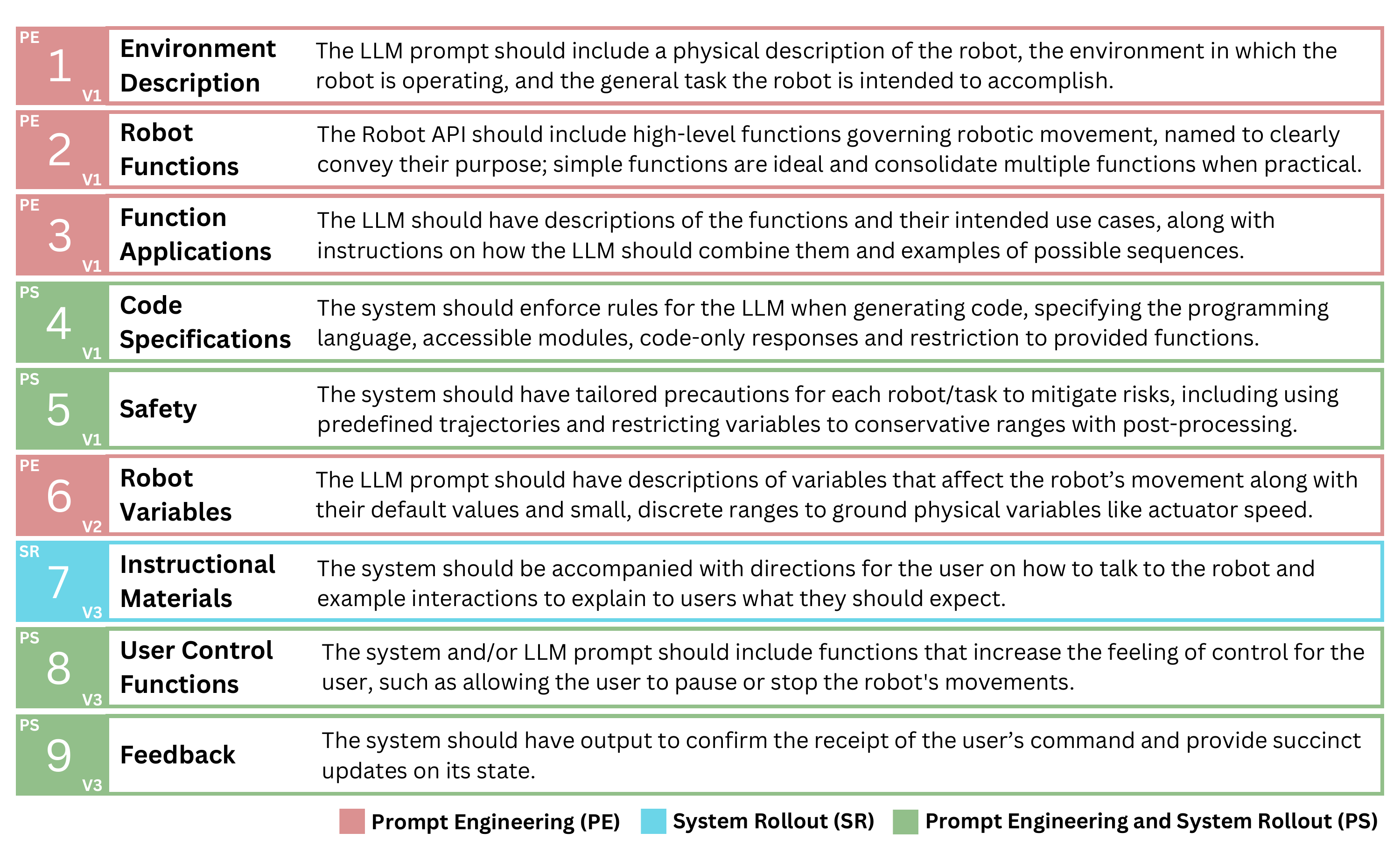}
  \caption{Our final framework consisting of 9 components is shown. The color and annotation at the top left of each numbered component indicates whether the component is related to prompt engineering (PE), system rollout (SR), or both prompt engineering and system rollout (PS). System rollout refers to the process of deploying the LLM-based speech interface and robot with users. The annotation at the bottom right of each numbered component shows which framework iteration it was added in.}
  \Description{Our final framework consisting of 9 components is shown. The color and annotation at the top left of each numbered component indicates whether the component is related to prompt engineering (PE), system rollout (SR), or both prompt engineering and system rollout (PS). The color and annotation at the bottom right of each numbered component indicates in which framework iteration it was added. The first 5 components: Environment Description, Robot Functions, Function Applications, Code Specifications, and Safety were added in Version 1. Robot Variables was added in Version 2. Lastly, Instructional Materials, User Control Functions, and Feedback were added in Version 3.}
  \label{fig:framework}
\end{figure*}

Our final framework, shown in Fig.~\ref{fig:framework}, and design guidelines, shown in Fig.~\ref{fig:design_guidelines}, provide insight on important human-centric considerations to researchers, engineers, and product designers for integrating off-the-shelf LLMs with assistive robots and devices. The contributions of this work are as follows: 
\begin{itemize}
    %\vspace{-0.5cm}
    \item We present an iteratively constructed framework, shown in Fig.~\ref{fig:framework}, for integrating LLMs as assistive speech interfaces for a robot. 
    \item We integrate an existing LLM with a commercial feeding robot using our framework and evaluate the interface through a human study with 11 older adults from an independent living facility.  
    \item We present a reflection and design guidelines, shown in Fig.~\ref{fig:design_guidelines}, based on the findings of the human study. 
\end{itemize}

\section{Related Work}
\label{relatedwork}

\subsection{LLMs for Robotics}

\change{
There have been numerous advancements in using LLMs for robotics~\cite{firoozi2023foundation, kawaharazuka2024real, vemprala2023grid, singh2023progprompt, liang2023code, vemprala2023chatgpt, brohan2023rt, zhang2023large, zhao2023chat, ahn2024autort, wang2024lami, mahadevan2024generative}. For the scope of this paper, we will focus on works that use off-the-shelf LLMs for high-level task planning through code generation, as these works are most pertinent to using LLMs as assistive interfaces.} \change{Liang et al. utilize codewriting LLMs with perception and control APIs to create reactive and waypoint-based trajectories for various robotic platforms, while Vemprala et al. develop design guidelines and a pipeline for integrating ChatGPT with different robots and simulation environments~\cite{liang2023code, vemprala2023chatgpt}. Simarily, Singh et al. and Arenas et al. provide prompting guidelines and methodologies for using LLMs in robotic tasks~\cite{singh2023progprompt, arenas2023prompt}. Other work has used ChatGPT as an interface for surgical robots and quadrupeds~\cite{pandya2023chatgpt, macdonald2024language}. These prior works primarily focus on high level planning and code generation which is just one aspect of designing a speech interface. Most utilize prompt engineering and evaluate their LLMs using accuracy, specifically if the LLM succeeds in generating the “correct” code that the researcher expects based on the provided prompt and command. While some papers include human studies to evaluate their methods, none evaluate their system with participants who interacted directly with the LLM-integrated robot. In comparison, our work includes multiple rounds of direct interaction between the robot and participants including with older adults who are close to the target population of individuals who need feeding assistance. This process allows us to identify human-centric elements essential to positive user perception of the system.  

Past works have also explored using LLMs with social and service robots~\cite{wu2023tidybot, hu2024deploying, kim2024understanding, kian2024can}. Wu et al.~\cite{wu2023tidybot} focus on learning personal preferences through language input for TidyBot, a cleaning robot, while Hu et al.~\cite{hu2024deploying} benchmark LLM performance for code generation for various service tasks. Both papers focus on how LLMs can be utilized to perform a specific task and similar to other papers, do not have participants interacting directly with the robot. The absence of these human studies limits the insights into improving the general interface design and human-robot interaction. Lastly, Kim et al. explore using LLMs with a social robot and focus on conversational tasks, such as having the robot guide the participant in making a drink~\cite{kim2024understanding}. The authors compare their robot against text and voice based agents and present design insights from a thematic analysis of the data obtained during the human study. Most of their findings are only applicable to social, conversational robots and do not generalize to physically assistive robots and tasks such as robot-assisted feeding. In comparison to all past work on using LLMs for robots, we iterate on our presented framework through multiple rounds of testing with users and we introduce design guidelines derived from our final human study. 
}

\subsection{Robot-Assisted Feeding}
\label{sec:speechinterfacefeeding}
Robot-assisted feeding systems employ robotic manipulators and perception systems to feed users either autonomously or semi-autonomously using input from users through assistive interfaces~\cite{park2018multimodal, gordon2024adaptable, jenamani2024feel, bhattacharjee2019community, belkhale2022balancing}. Voice control interfaces have been explored in past works on robot-assisted feeding, but use keyword detection in comparison to LLMs, which could enable more natural speech interactions with the robot~\cite{bhattacharjee2020more, nanavati2023design}. Bhattacharjee et al. find that participants prefer using speech over a web interface while dining alone~\cite{bhattacharjee2020more} while Nanavati et al. also find that participants see value in speech interfaces in quieter social settings~\cite{nanavati2023design}. In both studies, the researchers find that participants may not want to use speech in social dining settings as it may interfere in conversations. Lastly, Nanavati et al. find that participants want customizable feeding robots that can be tailored to specific needs and take into account their preferences~\cite{nanavati2023design}. These works motivate the need for the development of customizable speech interfaces for assistive feeding robots. 

Alongside research progress in assistive feeding, commercial feeding robots, such as the Obi feeding robot, are being used by hundreds of individuals across the United States. Obi, shown in Fig.~\ref{fig:obi_study3_setup}, is a FDA compliant Class 1 medical device from DESĪN LLC~\cite{Obi}. The robot consists of a 6 degree-of-freedom robot arm with a spoon end-effector and a dish tray consisting of 4 bowls. The commercial system uses hard-coded trajectories for the utensil to scoop food from a bowl and transfer that food to a predefined location near the user's mouth. For this work, we utilize a research version of the Obi robot with a Python API that allows us to send trajectories of joint angles to the robot. Obi is currently only commercially sold with a two button interface, with one button to select the bowl and the other button to scoop from the selected bowl. The button interface can be limiting especially for users who may want to customize their assistance; in comparison, an LLM-based speech interface can allow a user to provide verbal instructions to Obi, much like they would to a human caregiver, allowing them to customize how the robot feeds them.

%\section{Framework and Iterative Design}
%We provide a framework and design guidelines for integrating LLMs into assistive robotics, and demonstrate its effectiveness through an assistive feeding robot, Obi, shown in Fig.~\ref{fig:obi_study3_setup}. We develop the framework through a literature review and in-lab study. Subsequently, we apply this framework in a study involving 11 older adults and provide accompanying design guidelines.
 
% We first conducted a literature review to establish the foundations for our initial LLM integration within our robot. Subsequently, we conducted an in-lab study to assess the effectiveness of our initial design. Incorporating valuable user feedback, we refined our system and proceeded to conduct a subsequent user study involving 11 older adults at [location redacted for anonymous review]. Based on the insights from these studies, we present a set of design guidelines for using LLMs as assistive interfaces for robots.

\section{Version 1: Framework from Literature}
\subsection{Framework Development}
Prior work has shown that off-the-shelf LLMs can be used for high level planning and code generation for robots when provided with a well-designed prompt~\cite{vemprala2023chatgpt, singh2023progprompt, arenas2023prompt, vemprala2023grid}. We start our design process by drawing on this existing work on prompt engineering for LLMs for robotic control to develop an initial framework. 

Our Version 1 framework identifies five preliminary components, shown and described in Fig.~\ref{fig:framework}. We describe how these preliminary components were motivated from past literature:
\begin{itemize} %[leftmargin=1em]
  \renewcommand{\labelitemi}{\ding{226}}
    \item Environment Description: Vemprala et al. suggest to include information in the prompt about the environment in which the task is taking place and the goals and objectives of the task~\cite{vemprala2023chatgpt}; likewise, Arenas et al. recommend including an overview of the robot and the task at hand~\cite{arenas2023prompt}. We combine insights from these past works to define Environment Description as a physical description of the robot, the environment in which the robot is operating, and the general task the robot is intended to accomplish. 
    \item Robot Functions: Past papers on using LLMs for high-level task planning and code generation suggest defining and listing a set of robot functions, \change{named the Robot API}, in the prompt so that the LLM can access these function names and use them to control the robot~\cite{vemprala2023chatgpt, arenas2023prompt, singh2023progprompt}. Similar robot functions are required for LLM-based assistive speech interfaces. For example, for an assistive cleaning robot, you may include a function to perceive objects and a function to grasp an object which can be used in conjunction by an LLM to generate code to detect and pick up toys from the floor of a cluttered room. 
    \item Function Applications: Beyond listing the names of the robot functions, Vemprala et al. and Arenas et al. recommend including documentation of what each function does in the prompt~\cite{vemprala2023chatgpt, arenas2023prompt}. Both works also suggest providing solution examples to demonstrate to GPT how to use the functions to accurately respond to users commands; Arenas et al. especially emphasize the importance of examples because the custom control functions are not in an LLM's training set, so the model is unfamiliar with them~\cite{arenas2023prompt}. Therefore, we dedicate a component of our framework to specifying what each function does and providing examples of how they are intended to be used.
    \item Code Specifications: In order to use an off-the-shelf LLM for code generation for robots, past work has shown that the LLM must be prompted with general details of the programming environment, including what programming language should be used and what packages are accessible~\cite{vemprala2023chatgpt, arenas2023prompt, singh2023progprompt}.
    \item Safety: While integrating LLMs with assistive robots, safety needs to be evaluated on a case by case basis depending on the robot and the task at hand. Vemprala et al. suggest having a human in the loop to inspect the code that GPT generates, which isn't practical for an assistive interface that should respond immediately to the user~\cite{vemprala2023chatgpt}. Other options include control-barrier functions~\cite{rauscher2016constrained, cortez2019control} and testing extensively in simulation before deployment~\cite{vemprala2023grid, gualtieri2021emerging}. Due to the especially crucial role of safety in the context of physically assistive robots, we suggest using lightweight or compliant robots and predefined trajectories, thus barring GPT from modifying low-level joint angles or robot positions. 
\end{itemize}

\subsection{Implementation}
To validate our framework, we implemented an LLM-based speech interface for the Obi feeding robot. We chose to use the Obi as it is a commercially available assistive robot designed with safety considerations, namely a lightweight arm and a detachable spoon in case of collisions. Additionally, as discussed in Section~\ref{sec:speechinterfacefeeding}, prior studies on robot-assisted feeding indicate a preference for voice control interfaces in specific scenarios like individual dining, underscoring the necessity for such interface development.

Using our initial framework, we crafted a tailored prompt for GPT-3.5 Turbo, subsequently referred to as GPT, designed for the Obi feeding robot. Our prompt, shown in Fig.~\ref{fig:prompt_iteration}, integrated the five identified components from our Version 1 framework. 

\setlength{\emergencystretch}{.5em}\par
We addressed the components as follows: to implement the Environment Description, within our prompt, we included details on the task at hand, feeding the user, and the working environment, such as the foods in each of the bowls, \change{as the Obi robot does not have a built-in perception system}. We incorporated the Robot Functions and Function Applications components by providing the names of three high-level robotic control functions in the prompt, along with a short description of what each function did and how it should be used. These three functions moved the robot arm along hard-coded trajectories and were described as follows: \verb|obi.scoop_from_bowlno(bowlno)| moved the robotic arm to the specified bowl and scooped food from it; \verb|obi.move_to_mouth()| moved the robot to the user's mouth position; and \verb|obi.scrape_down_bowlno(bowlno)| moved the robot to the specified bowl and subsequently scraped food from the sides to the center of that bowl so that the robot could pick up additional food. \change{We designed the Robot API to include these three functions because they correspond to actions performed by the commercially available Obi robot. Specifically, the robot's default two-button interface allows users to select the bowl with one button and scoop/feed themselves with the other. Additionally, the scraping motion occurs automatically after every few scoops. }

At the end of the prompt, to satisfy the Code Specifications component, we included sentences instructing GPT to provide Python code for each request. Lastly, to address the Safety component, each function only moved the robot along a predefined trajectory, as opposed to allowing GPT to directly dictate the joint angles of Obi's arm. This prevents GPT from generating unsafe or unanticipated trajectories. Choosing to use the Obi robot, which is specifically designed to be safe for individuals with motor impairments, instead of other robotic feeding systems, is another implementation decision for enhancing safety. 

At this stage, users would hold down a physical button while speaking to record a command to the robot, which would be transcribed to text using OpenAI's Whisper API~\cite{whisper}. The parsed speech-to-text commands were appended to the bottom of the prompt and provided to GPT, which generated Python code that could subsequently be deployed on the Obi robot.

\subsection{Pilot Testing with Lab Members}
\change{We conducted pilot testing of our Version 1 interface for the Obi robot within our research team and members of our lab group. Four lab members watched videos of the robot performing four feeding tasks and suggested different command phrasings to accomplish the task, resulting in 48 phrases. These commands were tested for accuracy, with mistakes analyzed and prompts adjusted iteratively. Later, three researchers directly interacted with the robot to further assess and improve the system, focusing on user comfort during assisted feeding. More details on this process is included in Appendix~\ref{appendix:v1pilottesting}.}

\section{Version 2: Improved Accuracy and Customizability}

\subsection{Framework Iteration}
Based on our pilot testing, we noted instances where GPT made errors, enabling us to identify gaps or areas within the framework that required iteration. \change{Specifically, the Code Specifications were not detailed enough, causing GPT to import nonexistent modules and define unnecessary new functions. To address this, the framework was modified to emphasize the importance of explicitly outlining how GPT should write code. Additionally, the Function Applications component was expanded to better describe the relationships between functions and provide examples, while the Robot Functions component was modified to suggest merging common sequential functions to improve performance in tasks that required multiple robot actions. More information on these changes can be found in Appendix~\ref{appendix:v2frameworkiter}.}

Finally, we added a new field to the existing framework:
\begin{itemize} %[leftmargin=1em]
  \renewcommand{\labelitemi}{\ding{226}}
    \item Robot Variables: Users wanted to have more ability to customize the way in which the robot moved. In the case of the Obi robot, the first iteration had only a single speed and scoop depth, yet, based on user feedback, we found that preferences for these can vary significantly from user to user, between different foods, and even from situation to situation. Accordingly, we added a component to the framework providing descriptions of variables affecting the way in which the robot moves, their ranges, and their default values. 
\end{itemize} 

Due to the user safety implications of the new Robot Variables component which gives the LLM speech interface control of motion variables such as speed, we proactively modified the Safety component of the framework to suggest restriction of robot variables such as speed to conservative ranges \change{and to check the validity of robot variables in LLM-outputted code to ensure they are within expected bounds.}

\begin{figure*}[hbt!]
  \centering
  \includegraphics[width = \textwidth]{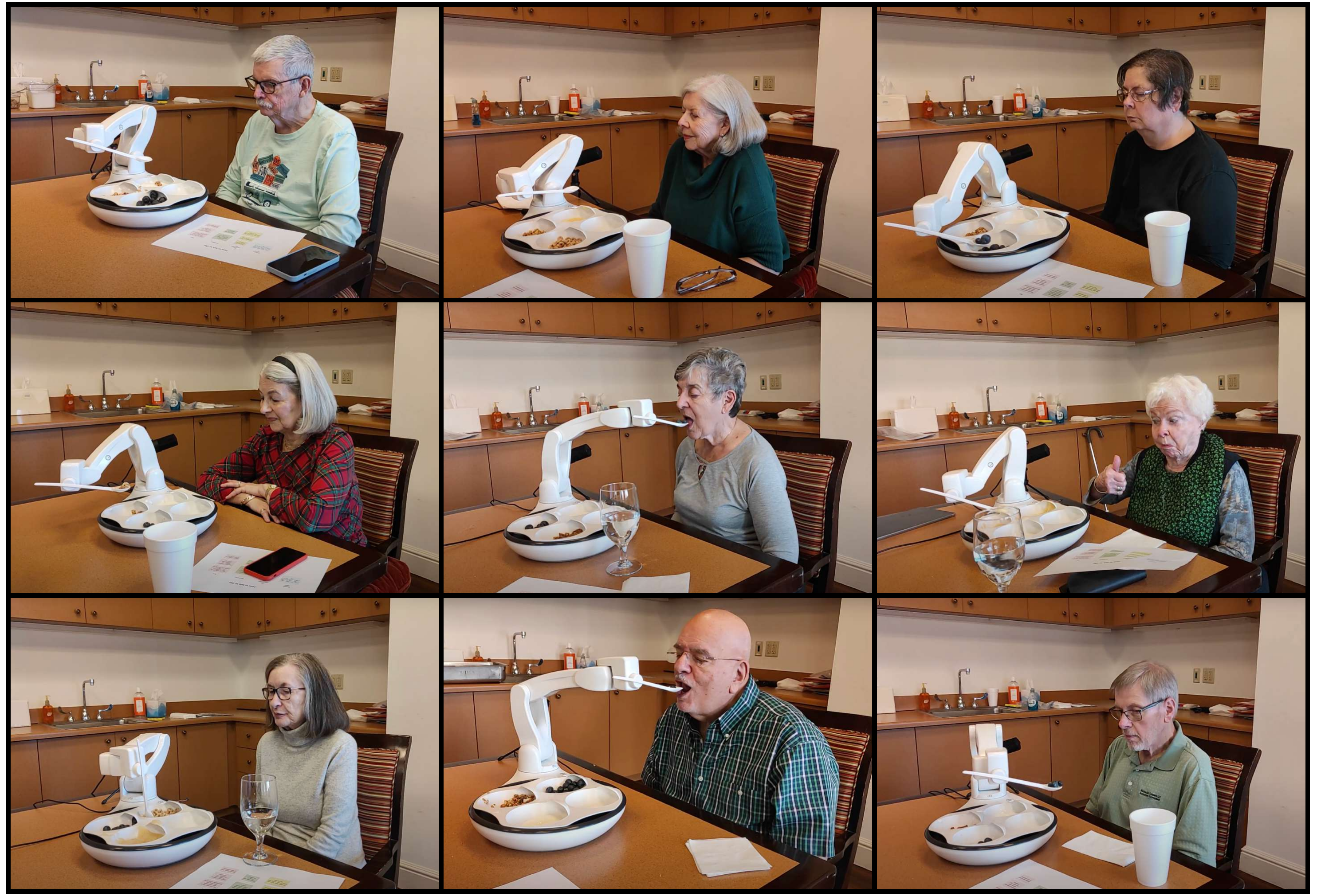}
  \caption{9 of the 11 participants are shown at various stages of the Version 3 study at an independent living facility with older adults.}
  \Description{9 of the 11 older adults from our study at an independent living facility are shown. In the clips, Obi is either scooping food, Obi is moving to the mouth, or the participant is taking a bite of food.}
  \label{fig:study3_compilation}
\end{figure*}

\subsection{Implementation}

At this stage, we changed the prompt for the Obi system to reflect the modifications to the framework and thus improve the accuracy of GPT's responses. Our updated prompt can be found in Fig.~\ref{fig:prompt_iteration}.

\change{In summary, we updated the prompt to emphasize the production of responses exclusively in Python code, without importing extra modules or creating new functions, as specified in the revised Code Specifications component. To adhere to the updated Function Applications component, we added detailed instructions on how the robot functions should be combined, emphasizing Obi's movement to the user's mouth when asked to be fed, and included a 4-second default pause between bites for better user experience, with customization options for the delay. Additionally, the Robot Functions component was modified to ensure that after scraping down a bowl, Obi would immediately scoop from the same bowl to improve accuracy, addressing previous issues where the sequence was misunderstood. For the Robot Variables component, we added new variables for Obi's movements (speed, acceleration, and scoop size) to enhance customizability, with post-processing by the system to ensure safety. More information on these changes can be found in Appendix~\ref{appendix:v2implementation}.}

\subsection{Demonstration with Community Members}
After the modifications to the implementation in Version 2, we brought the system to a disability awareness event in Pittsburgh, PA, USA, where we demonstrated the speech interface to about 15 members of the community, some of whom had motor impairments or other types of disabilities. During the event, community members who wanted to interact with the robot provided commands to the feeding robot to have it feed a lab member either M\&Ms or Cheerios cereal placed in two bowls of the Obi. We encouraged visitors to be creative with their requests to the robot. Most individuals who interacted with the robot provided only 1-2 commands and the interactions were short, lasting only a few minutes. Nevertheless, this process was helpful for evaluating the system, enabling us to identify where users were confused or struggled to use the interface, as well as hear feedback on what people liked and disliked. 

\section{Version 3: Increased Ease of Use}
\subsection{Framework Iteration}
Using insights from the Version 2 demonstration, we identified essential modifications and new components for our framework. \change{First, we made adjustments to the Robot Variables component due to GPT's inconsistent handling of continuous variables, which caused unpredictable speed changes that confused users. By grounding variables like actuator speed on a 0-5 scale, speed adjustments became more consistent and aligned with user expectations. This process allowed us to identify that while increasing the granularity of robot function and variables can enhance the user perception of robot customization, they can negatively impact the language model's accuracy. More details are included in Appendix~\ref{appendix:v3frameworkiter}.}

Additionally, we added three new components to the framework, all of which focus on human-robot interaction: 
\begin{itemize} %[leftmargin=1em]
  \renewcommand{\labelitemi}{\ding{226}}
\item Instructional Materials: Users tended to struggle with giving commands to the system at first, as they were unsure what they were allowed to say or what the robot was capable of and were wary of experimenting with new commands. Thus, it became evident that the framework required a field on how to effectively prepare users on how to use a speech interface by providing guidelines for user communication with the robot and sample interactions to set their expectations. This component was added to emphasize the importance of giving users more guidance and support with using the system beforehand to reduce user confusion.
\item User Control Functions: A component was added to incorporate functions that increase the feeling of control for the user, an important consideration for assistive interfaces~\cite{padmanabha2024independence, bhattacharjee2020more, javdani2018shared}. Example functions include allowing the user to pause or stop the robot’s movements using voice commands.
\item Feedback: Users struggled to discern whether the system heard them or not, especially given the few second delay between completion of the command and robot movement due to networking and GPT processing the request. As a result, they would sometimes attempt to repeat themselves or become confused or frustrated. Thus, we added a Feedback component to the framework to convey the need for feedback from the system to the users informing them about the current state of processing and execution and providing a clear error message if it is unable to process their command correctly. 
\end{itemize}

\change{In comparison to all related work, these components move beyond governing only prompt engineering, instead providing recommendations for designing the system as a whole in a user-centric way.}

\subsection{Implementation}
We updated the implementation for the Obi robot using our latest framework, aiming to enhance user acceptance of the system. Our updated prompt is shown in Fig.~\ref{fig:prompt_iteration}. 

\change{First, to adhere to the updated Robot Variables component on grounding of physical variables, we changed the continuous speed and acceleration ranges provided to GPT from [0, 80] degrees per second and [0, 250] degrees per second squared respectively to discrete [0, 5] ranges. Next, for the Feedback component, we changed how users interact with the system by replacing the need to hold down a button with a wakeup phrase, "Hey Obi," implemented via the Porcupine Python API from PicoVoice. We also added audio cues to inform users when the wakeup phrase is registered, when their request is being processed, and when the robot is about to execute or has completed an action. For the Instructional Materials component, we created an instructional video and a reference sheet, included as Supporting File 1 and 2, to help users understand and use the system effectively. These materials are expected to be temporary aids as users become familiar with the robot. More details can be found in Appendix~\ref{appendix:v3implementation}.}

Finally, we implemented three new User Control Functions that GPT is able to call, giving users the ability to interrupt and halt the movement of the robot with a verbal command for any reason: 
\verb|obi.start()|, which begins or resumes execution of any robot code;
\verb|obi.stop()|, which permanently stops execution of any currently running robot code; and
\verb|obi.pause_indefinitely()|, which suspends execution of any currently running robot code. 

\subsection{Testing with Older Adults}
To evaluate Version 3 of the implementation, we conducted a formal study at an independent living facility, Baptist Providence Point in the greater Pittsburgh, Pennsylvania metropolitan area, with 11 \change{non-disabled} older adults (5M, 6F) ranging in age from 72-91 (mean = 81.1, SD = 5.9). A subset of the participants are shown in Fig.~\ref{fig:study3_compilation}. The ethnicity of all participants was white. Older adults were chosen in the study as they share more similarities with the target demographic of individuals with physical impairments as compared to the prior populations the interface was tested with due to their age. Our study, approved by Carnegie Mellon University's Institutional Review Board, involved the researchers traveling to the independent living facility and conducting one-hour-long studies with each participant. Recruitment was conducted in person through an information session held at the independent living facility as well as over email, and informed consent was obtained from the participants for use of image in publication. Participants were asked how much experience they have with controlling a robot on a 7-point scale ranging from ``1 = no experience'' to ``7 = expert user'' and how they feel about using robots to help with everyday tasks ranging from ``1 = very negatively'' to ``7 = very positively''. Participants had a median experience of 1 (IQR = 1, min = 1, max = 4) and a median attitude of 6 (IQR = 1, min = 4, max = 7). 

As seen in Fig~\ref{fig:obi_study3_setup}, we placed the Obi robot and a microphone directly in front of the participant. If needed, we adjusted the food delivery location of the robot to a comfortable location for the participant directly in front of their mouth. As discussed previously, in line with the ``Instructional Materials'' field of our framework, we started the study by showing the participant an instructional video and providing them with a reference sheet---both included as Supporting Files. 

After the instructional session, the participant was asked to select three foods out of the following options: blueberries, yogurt, granola, pudding, pretzels, and Cheerios. As seen in Fig~\ref{fig:obi_study3_setup}, these foods were added to bowls 0, 1, and 2 while bowl 3 is left empty. The first part of the study consisted of 1 practice task, eat a single scoop from any bowl, and 5 predefined tasks: (1) eat a single scoop from bowl 0 quickly, (2) eat a single large scoop from bowl 2, (3) scrape bowl 0 and eat a single scoop from bowl 0, (4) eat three scoops from bowl 1, (5) eat a single scoop from bowl 2 followed by single scoop from bowl 0. For each task, participants were instructed to only provide a single command. After each task, participants were asked a yes/no question, ``Did the robot adequately complete the intended task?'' If the participant answered no, they were given an additional attempt to complete the task, up to 3 total attempts. \change{For each task, interaction memory, the previous commands given by the user to the robot and the previous outputs by GPT, was provided to GPT for new attempts.} All surveys and questionnaires asked to participants during the study are included in the Appendix~\ref{appendix:questionnaires}.

%After the task, participants are asked 3 7-point Likert items: ``I found it easy and straightforward to accomplish the specified task,'' ``I didn't have to put in any mental effort to complete the task,'' and ``I did not feel frustrated while completing the task.'' 

The next part of the study was a 7 minute open feeding session. If needed, we refilled the bowls with more food. The instructions provided to the participants were ``Your goal is to finish the meal in a pleasant and efficient way. You are free to command the robot to eat the food however you like. Feel free to be creative in how you accomplish this goal.'' \change{For this part of the study, interaction memory was provided to GPT for the entire session.} 

At the end of the study, we asked participants for responses to the NASA TLX workload items on a 0-100 scale, the System Usability Scale (SUS), multiple 7-point Likert items shown in Appendix~\ref{appendix:likert}, and 4 open ended questions included in Appendix~\ref{appendix:open-ended}. We also audio and video recorded the entire study to capture participant reactions, feedback, and responses to the open-ended questions.

\begin{figure}[t!]
  \centering
  \includegraphics[width = \columnwidth]{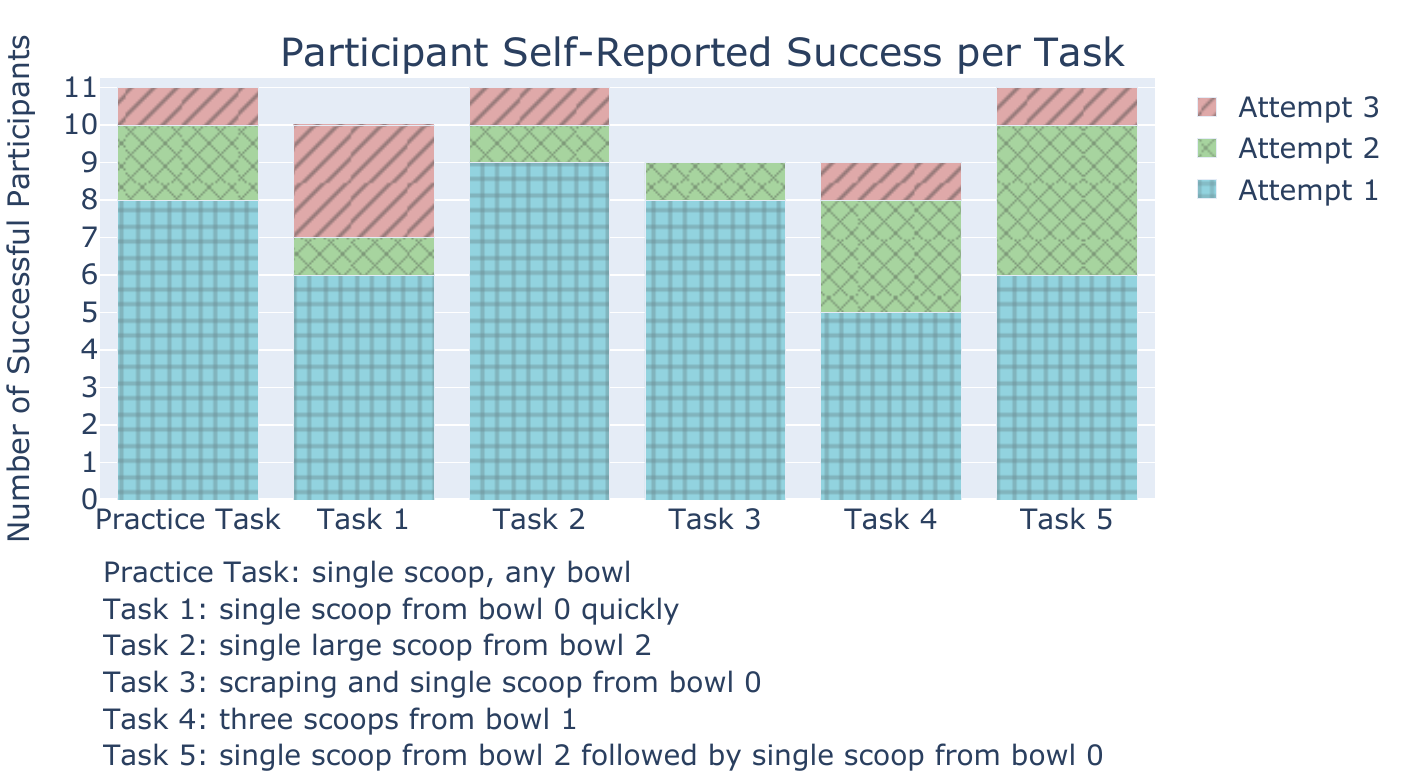}
  \caption{Participant self-reported success per task for each attempt. After both the practice task and the 5 predefined tasks, participants reply Yes/No to the question ``Did the robot adequately complete the intended task?''}
  \Description{The number of successful participants for each task is shown as a bar chart. Each bar conveys at which attempt (1,2, or 3) the participant was successful. }
  \label{fig:attempts}
\end{figure}

\begin{figure}[t!]
  \centering
  \includegraphics[width = \columnwidth]{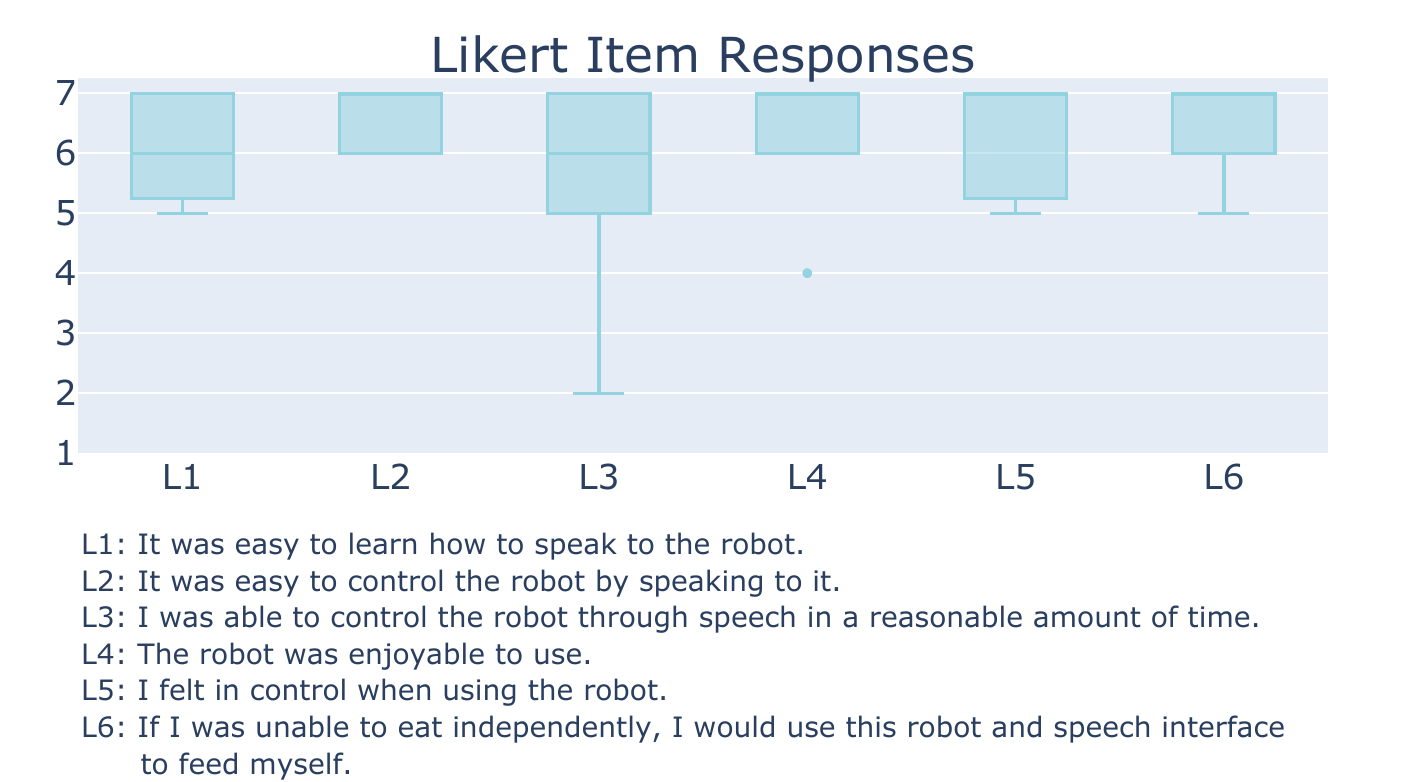}
  \caption{Responses from six 7-point Likert Items answered by the 11 participants at the end of the study. A high score is best for all items, with 1 = Strongly Disagree and 7 = Strongly Agree.}
  \Description{Data from the six 7-point Likert Items are shown. L1 is It was easy to learn how to speak to the robot. L2 is It was easy to control the robot by speaking to it. L3 is I was able to control the robot through speech in a reasonable amount of time. L4 is The robot was enjoyable to use. L5 is I felt in control when using the robot. L6 is If I was unable to eat independently, I would use this robot and speech interface to feed myself. The median responses for the six items respectively are 6 (Agree), 7 (Strongly Agree), 6 (Agree), 7 (Strongly Agree), 7 (Strongly Agree), and 7 (Strongly Agree).}
  \label{fig:Likert_Scale}
\end{figure}

\change{
Both qualitative and quantitative data from our study show that our framework and methodology were successful in creating a positively-received speech interface for the Obi feeding robot. As seen in Fig.~\ref{fig:attempts}, the majority of participants were able to complete the predefined tasks within 3 attempts, showing adequate accuracy for the LLM interface. \change{For reference, we include all participant commands in Appendix~\ref{appendix:participant_commands}.} From the responses to Likert items shown in Fig.~\ref{fig:Likert_Scale}, we find that participants found the interface easy to learn, answering with a median score of 6 (Agree) to L1 (Likert Item 1), ``It was easy to learn how to speak to the robot'', with one participant (P10) saying, ``I would imagine that most people would learn to use the system quickly.'' They also found the robot easy to control through speech and felt in control when using the robot with median scores of 7 (Strongly Agree) for both items L2 (Likert Item 2) and L5 (Likert Item 5). P5 summarized this at the end of the study saying, ``What I liked was it gave me a sense of control. For somebody who's in an incapacitated situation, that would be very important... And it was easy to talk to.'' Lastly, participants also gave a strong median rating of 7 (Strongly Agree) for items L4 (Likert Item 4) and L6 (Likert Item 6), related to enjoyment while using the robot and if they would use the robot and speech interface if they were unable to eat independently. 

Using the NASA TLX scale, we find that participants generally reported low workload measures across all categories as seen in Fig.~\ref{fig:NASA_TLX}. Lastly, the average System Usability Scale score was 73.0 (SD = 18.6, min = 50.0, max = 95.0), which shows decent performance, but leaves room for improvement. The strong Likert item responses, workload measures, and SUS scores from older adults show promise for using LLMs as assistive interfaces for robot-assisted feeding and other physical tasks. Future testing with more individuals, especially those unable to feed themselves, is needed for full validation of the system. To aid researchers in robot-assisted feeding, we additionally include bite-timing metrics in Appendix~\ref{appendix:bite_timing_metrics}.
}

\section{Reflection}

\subsection{Final Framework}
\change{We present our final framework for integrating LLMs as assistive speech interfaces, outlined and detailed in Figure \ref{fig:framework}. The framework now comprises a total of nine components: Environment Description, Robot Functions, Function Applications, Code Specifications, Safety, Robot Variables, Instructional Materials, User Control Functions, and Feedback. In the figure, we indicate if each component pertains to prompt engineering (PE), system rollout (SR), or both prompt engineering and system rollout (PS).}

\subsection{Analysis}
To gain deeper insights into the needs of users of LLM-based speech interfaces for robots, we conducted a thematic analysis on audio recordings collected during our study with older adults. Thematic analysis is a common technique used for analyzing qualitative feedback from participants in human-robot interaction studies~\cite{kim2024understanding, soraa2023older, rogers2020defining, cresswell2018health}. This analysis informed the development of design guidelines for integrating LLMs as speech interfaces for assistive robots. While our framework provides assistance with integrating an LLM as a speech interface, our design guidelines should be strongly considered during development and testing of such interfaces as they can further enhance the user experience. 

Initially, we transcribed the audio recordings from our user studies. Subsequently, we uploaded these transcripts into ATLAS.ti~\cite{atlas} and employed deductive coding to extract users' perceptions of the system. Our aim was to comprehend both the effective and challenging aspects of our system; therefore, we coded user feedback into three distinct categories: (1) negative responses to the interface, (2) positive responses to the interface, and (3) desired features.

After coding the transcripts using the defined categories, we proceeded to analyze the quotes within each category using affinity diagramming. This process allowed us to identify and group similar feedback provided by users. We conducted two iterations of coding and affinity diagramming to ensure that we captured all relevant data comprehensively. During this iterative process, we focused on identifying challenges and needs expressed by users and grouped similar feedback together. By examining the collective feedback, we synthesized our findings into actionable guidelines provided below.

% Then, an affinity diagram was created for each of the categories where quotes from that specific category were grouped together based on a general problem trend that they indicated. This process was repeated once more to ensure that we no quotes were missed and a separate affinity diagram was created for them. Finally, the diagrams from the two different iterations were combined to obtain the final set of guidelines that we present in this paper.

\begin{figure}[t!]
  \centering
  \includegraphics[width = \columnwidth]{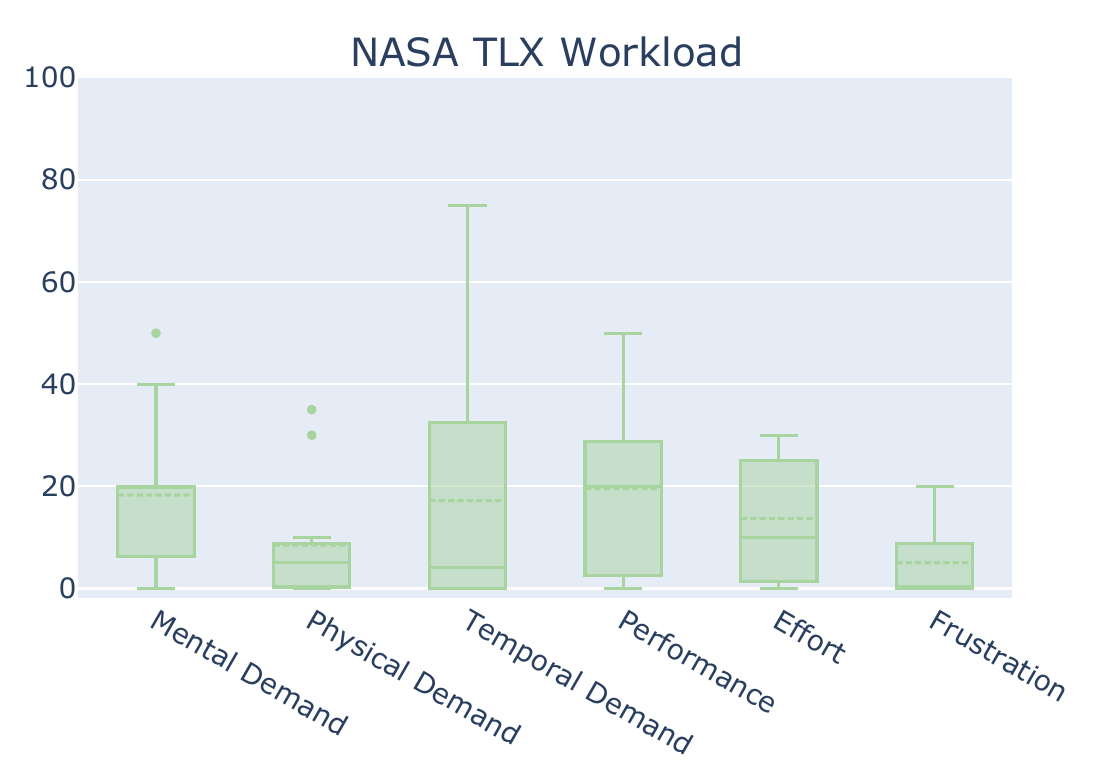}
  \caption{Responses to a 0-100 NASA TLX Scale assessed at the end of the study with all 11 participants. A low rating is best for all categories.}
  \Description{Data from the 0-100 NASA TLX Scale is shown for the six categories: Mental Demand, Physical Demand, Temporal Demand, Performance, Effort, and Frustration.}
  \label{fig:NASA_TLX}
\end{figure}

\begin{figure*}[hbt!]
  \centering
  \includegraphics[width = \textwidth]{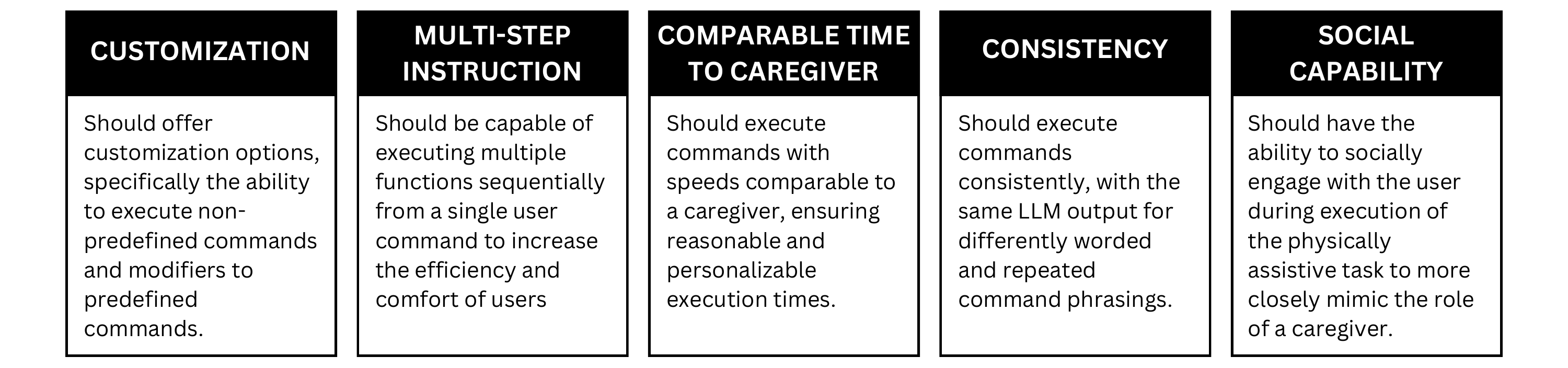}
  \caption{Design Guidelines. Using both quantitative and qualitative data from a human study with 11 older adults, we establish 5 guidelines for integrating LLMs as speech interfaces for physically assistive robots.}
  \Description{The 5 design guidelines: Customization, Multi-Step Instruction, Comparable Time to Caregiver, Consistency, and Social Capability are shown with brief descriptions.}
  \label{fig:design_guidelines}
\end{figure*}

\subsection{Design Guidelines}
Based on our analysis of user feedback from the study at an independent living facility, we provide design guidelines, shown in Fig.~\ref{fig:design_guidelines}, for the incorporation of LLMs in assistive interfaces. The recommendations include (1) Customization (2) Multi-Step Instruction, (3) Consistency, (4) Comparable Time to Caregiver and (5) Social Capability. We discuss each guideline in more detail below.

\subsubsection{Customization}
\begin{figure*}[hbt!]
  \centering
  \includegraphics[width = \textwidth]{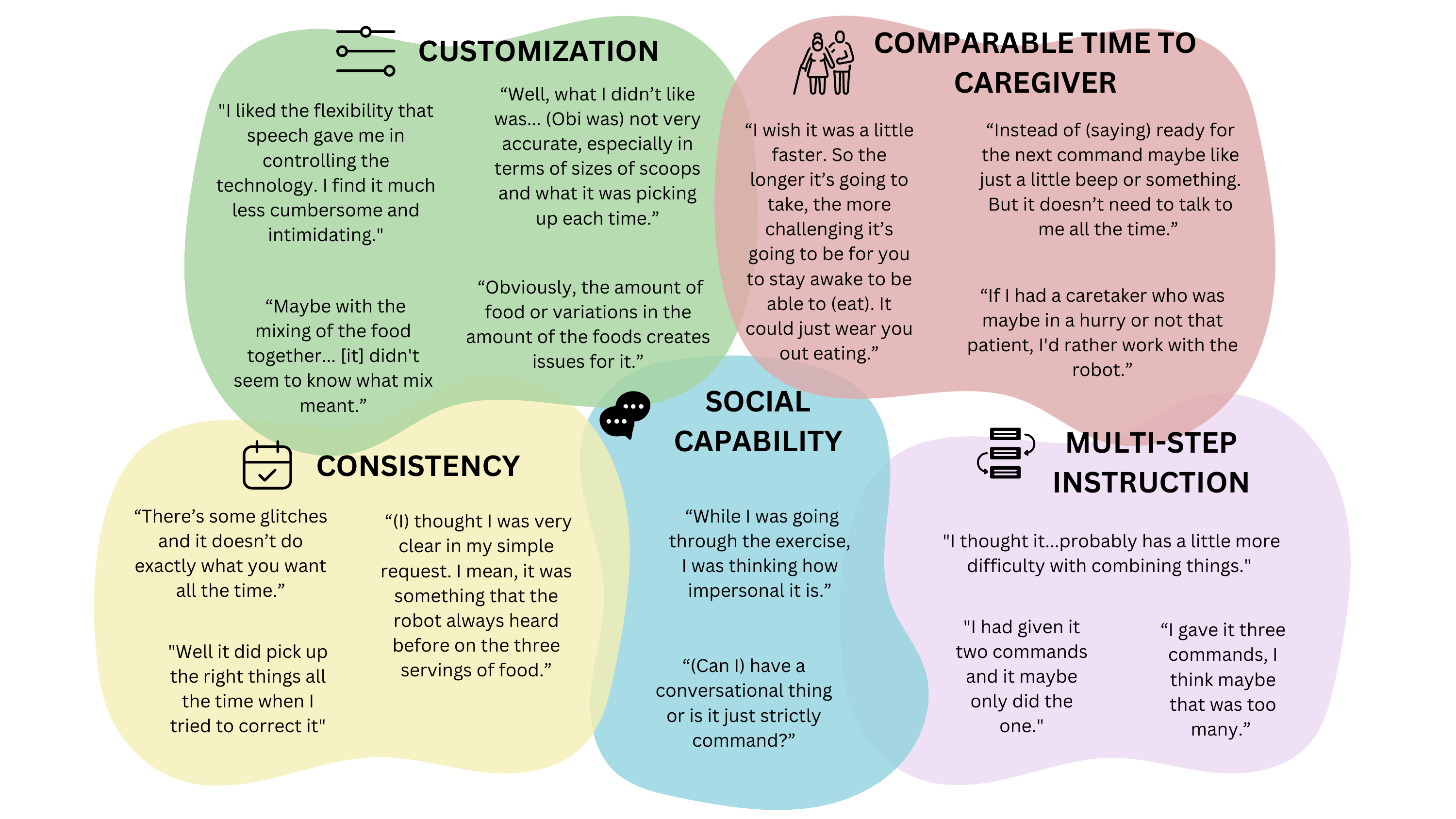}
  \caption{A subset of quotes from our study involving 11 older adults is presented and organized according to the categories outlined in our design guidelines.}
  \Description{A subset of quotes from our study involving 11 older adults is presented and organized according to the categories outlined in our design guidelines.}
  \label{fig:guidelines_quotes}
\end{figure*}

Customization refers to the ability of the user to incorporate their preferences while using an interface to complete an assistive task. All 11 participants provided feedback and comments regarding the customization of the robot through the interface, with select quotes from participants depicted in Fig.~\ref{fig:guidelines_quotes}. Through our thematic analysis, we find that the ability for users to provide Obi commands customized to their own preferences (such as how big of a scoop they wanted) was pivotal for successful interactions. This design guideline directly aligns with insights from past works on the importance of user customization of robotic interfaces~\cite{ranganeni2024customizing, ranganeni2023evaluating, padmanabha2024independence}. From our analysis, there are two types of ways that LLM-based speech interfaces must be able to provide users with customization: (1) Modifiers and (2) Ability to process non-predefined commands.

    \textbf{Modifiers:} Modifiers refer to an attempt by the user to customize the execution of commands, using words like ``faster,'' ``slower,'' ``larger,'' and ``smaller.'' For example, for the Obi robot, this might involve a desire for the robot to move at a faster speed while moving to the mouth. Throughout our study, both for the predefined tasks and the open feeding session, participants frequently issued commands to the robot accompanied by modifiers, but some expressed concerns with the system's ability to consistently understand their modifiers. 

    The quantitative data, shown in Fig.~\ref{fig:attempts}, specifically Task 1 and 2 which involved increasing robot speed and scoop size, highlights some of these challenges that participants faced with modifiers. For Task 1, only 6 participants were successful in one attempt, which shows room for improvement. P10 requested Obi to increase the speed during feeding, noting that ``the change in speed seemed to be not great.'' Similarly, P2 highlighted issues when asking for different bite sizes saying, ``what I didn't like was... [Obi was] not very accurate, especially in terms of sizes of scoops and what it was picking up each time.'' Specifically, 6 participants expressed challenges with modifying the scoop size. We would like to note that scoop size challenges were primarily due to the lack of a perception system to dynamically change the depth of scoop based on the amount of food in the bowl. Our system instead used a naive approach and included a Robot Variable named ``scoop depth'' in the prompt that could be modified by the LLM to adjust the depth of the scoop but didn't take into account the amount of food left in the bowl, leading to inconsistency in scoop sizes. 

    Nevertheless, the importance of modifier customization was underscored by participant quotes that unveiled frustration and confusion when the system didn't respond properly to their modifiers. For example, P5 expressed frustration with changing the scoop size saying, ``Trying to get the amounts is a challenge.'' We want to emphasize that modifiers are personalized, as they may carry varying interpretations for different users. Thus, the system should ideally adjust its execution to align with the user's expectations. Possible approaches to achieve this include accurate sensing and perception of the environment and learning user preferences, with past work demonstrating LLMs are well suited for this task~\cite{wu2023tidybot}.

    \textbf{Ability to process non-predefined commands:} Predefined commands refer to the Robot Functions, Function Applications, and User Control Function components of our framework; for the Obi robot, these commands include scooping from a bowl, moving to the mouth, etc. On the other hand, non-predefined commands denote additional functionalities that users may want, such as mixing for the Obi robot, which are not explicitly defined and described to the LLM. Non-predefined commands are essential to an assistive interface because they allow for increased customization of robot motions in addition to modifiers, allowing the robot to more closely adhere to the user's preferences. However, having an LLM understand and execute on non-predefined commands is challenging, especially while hard-coding robot trajectories in alignment with the Safety component of our framework. This challenge was apparent in our study, with 6 participants expressing difficulties in executing non-predefined commands, depending on the specific request and phrasing of the command. Some common non-predefined commands from the open-feeding session with Obi included mixing of foods and selection of a specific quantity of a food, such as 2 blueberries in one scoop. 
    
    While the vast majority of speech interfaces for assistive robots are unable to handle non-predefined commands, an advantage of LLMs is they have the ability to sequentially arrange predefined commands, which serve as `primitives,' to perform non-predefined commands. In the case of the Obi, during our human study, 5 participants requested the robot to mix foods, a non-predefined command. For 2 participants, GPT was successful, as it consecutively ordered scooping functions from each bowl leading to a bite with multiple foods as seen in the sequence of images in Fig.~\ref{fig:teaser}. However, the inability to execute non-predefined commands for some participants led to confusion. For example, P1 remarked ``Maybe with the mixing of the food together... [it] didn't seem to know what mix meant.''

\textbf{Summary:} Both predefined/non-predefined commands and modifiers aren't exclusive to the Obi robot; interfaces for other assistive robots, such as mobile manipulators, often contain predefined commands such as autonomous grasping of an object and modifiers such as the ability to change robot speed. Our thematic analysis showed that participants wanted the ability to customize their assistance using both modifiers and non-predefined commands. When these requests were not fulfilled, this led to frustration and confusion for users, underscoring the importance of incorporating these customization considerations while designing LLM-based speech interfaces. In contrast, when the LLM-based interface worked well for customization, users expressed strong approval; for example, P10 said ``I liked the flexibility that speech gave me in controlling the technology. I find it much less cumbersome and intimidating.'' 

\subsubsection{Multi-Step Instruction}
Multi-Step Instruction refers to the capability of an LLM-based speech interface to execute sequential robot actions encompassed within a single user command to the robot. This guideline is closely tied to the Function Applications component of our framework in which we discuss how to prompt an LLM to understand how to combine robot functions sequentially. From our thematic analysis, we find that multi-step instructions are important to users as they can help increase system acceptance by reducing frustration and user workload. 5 participants commented on multi-step instructions, with a subset of quotes shown in Fig.~\ref{fig:guidelines_quotes}. Some examples of multi-step instructions for the Obi robot are predefined Task 4, eating three scoops from bowl 1, and predefined Task 5, eating a single scoop from bowl 2 followed by a single scoop from bowl 0. As seen in Fig.~\ref{fig:attempts}, the system worked adequately for the respective tasks, with 9/11 and 11/11 participants completing the task within 3 attempts. However, there is still additional room for improvement as the system didn't work for a subset of users during their initial attempts. P3 summarized their experience, saying ``I thought it... probably has a little more difficulty with combining things'' while P7 stated, ``I gave it three commands, I think maybe that was too many.'' Consistent functionality of multi-step instructions is crucial to consider when using LLM-based speech interfaces for assistive robots as they can increase the efficiency and comfort of users, while reducing fatigue and frustration. Multi-step instructions also enable users to conduct activities such as watching TV or conversing, while having confidence that the robot will continue the task without additional input for an extended period of time.

\subsubsection{Comparable Time to Caregiver}
Utilizing assistive interfaces to control robots often results in individuals with impairments taking substantially more time to complete tasks in comparison to \change{non-disabled} individuals and caregivers~\cite{padmanabha2024independence, robotsforhumanity}. LLM-based speech interfaces for robots should strive to execute commands with speeds comparable to a caregiver, thereby enhancing user comfort and satisfaction. In total, 4 participants provided feedback on different aspects of the robot's execution time, with a subset of quotes shown in Fig.~\ref{fig:guidelines_quotes}. The robot's execution time encompasses networking and code generation time by the LLM, the duration for the robot to move and carry out the command, and the time for the robot to provide feedback to the user on its state. P10 commented on the feedback time, saying ``Instead of (saying) `ready for the next command' maybe like just a little beep or something. But it doesn't need to talk to me all the time.'' This insight ties directly to the Feedback component of our framework, and users of our framework should be mindful of providing feedback to users as succinctly as possible. In comparison, P1 highlighted the time taken for the entire feeding process, saying ``I wish it was a little faster. So the longer it’s going to take, the more challenging it’s going to be for you to stay awake to be able to [eat]. It could just wear you out eating.'' P1's comment highlights the impact of slow execution times on user fatigue, emphasizing the need for interfaces to align with the personalized time-frames typically provided by caregivers. In contrast to the 4 participants who felt the feeding process was too slow, P11 commented on their preference for the system over a rushed caretaker, saying, ``If I had a caretaker who was maybe in a hurry or not that patient, I'd rather work with the robot.''

Additionally, as seen in Fig.~\ref{fig:Likert_Scale}, participants responses to L3 (Likert Item 3), related to controlling the robot in a reasonable amount of time, had high variance, indicating that preferences regarding time-frames are strongly individualized. This observation is also evident in participant responses to temporal workload on the NASA TLX scale, shown in Fig.~\ref{fig:NASA_TLX}. 

Our findings indicate that executing commands within a similar time-frame as a caregiver would help improve user experience with LLM-based speech interfaces. This can be accomplished by giving precedence to preventing fatigue caused by temporal demand, identifying the specific moments during robot execution when users are prone to experiencing it, and streamlining or allowing customization to the level of communication needed between the user and the interface.

\subsubsection{Consistency}
\label{consistency}
Consistency refers to the ability of an LLM-based speech interface to reliably generate and execute commands. Specifically, these interfaces should produce the same output when given identically phrased commands repeatedly, and when given commands with similar intention, but that are differently worded. 6 participants commented on the system's limitations in reliably executing commands as expected. Some of these quotes are shown in Fig.~\ref{fig:guidelines_quotes}. For example, P4 said, ``(I) thought I was very clear in my simple request. I mean, it was something that the robot always heard before on the three servings of food,'' suggesting that Obi performed differently when executing the same command on different occasions. Despite participants encountering occasional consistency issues, we found that they were often able to rectify the behavior of the LLM. P2 remarked, ``Well it did pick up the right things all the time when I tried to correct it.'' 

A possible solution for improving LLM consistency is to modify parameters such as the temperature parameter for GPT-3.5 Turbo, which defines the randomness of LLM responses. In Section~\ref{sec:limitations}, we also discuss how LLM fine-tuning can be a possible remedy. Nevertheless, from the perspective of the user, it is essential that these interfaces consistently execute commands, as this builds trust and improves the system's usability. We recommend that LLM-based speech interfaces should be evaluated extensively for consistency before deployment.

\subsubsection{Social Capability}
Social capability pertains to allowing a user to interact socially with a robot while receiving physical assistance, similar to interactions with a human caregiver. 3 out of 11 participants provided insights, with some of the comments showcased in Fig.~\ref{fig:guidelines_quotes}, regarding the integration of social capabilities into the Obi system. P11 said ``While I was going through the exercise, I was thinking how impersonal it is.'' They further elaborated that they were unsure about whether they would prefer to be fed by Obi or a caretaker because of its lack of conversational abilities.  While none of the participants attempted to converse with Obi, P8 inquired about it, saying ``(Can I) have a conversational thing or is it strictly command?'' The feedback from users demonstrates the significance of integrating social capabilities, even though such capabilities are often not the primary focus of physically assistive robots. The presented social capability guideline is in agreement with past work on using LLMs as speech interfaces for social robots that shows that robot's social aspects can enhance user engagement~\cite{kim2024understanding}. We encourage researchers to explore the social capabilities of LLM-based speech interfaces for physically assistive robots, as this could allow the robot to provide a sense of companionship and engagement similar to that of a caregiver, thereby potentially increasing user acceptance.

%\subsubsection{Other Insights}
%Some participants were unsure if they enjoyed using the interface while some found it really intuitive. For example, P1 said that ``I think that, you know, people that are of my age are not used to using voice commands'' implying that older adults may not all be used to talking to their devices. P4 also was uncertain and said ``Is there something wrong that I'm saying'' several times during the study implying that they weren't completely comfortable with the nuances of using a speech-based interface. This suggests that there is room for multi-modality to make the system more intuitive and easy to use. While this doesn't mean that we either give users a completely speech-based or non speech-based interface to control Obi with, it could mean giving them the ability to perform different actions with different forms of input (like waking it up with a button but asking it to perform specific feeding tasks with speech commands.)

\subsection{Considerations and Future Work}
\label{sec:limitations}
In addition to the testing already conducted, a follow-up study involving individuals who are unable to feed themselves, \change{including long-term deployments}, would be beneficial to our framework and design guidelines. \change{These additional studies will enable us to determine if the LLM interface in conjunction with the robot meets specific needs and desires of users.} Furthermore, it would be useful to conduct our framework iteration process with another assistive robot such as a mobile manipulator that can do multiple tasks in the home. \change{This testing could present unique insights that users of our framework may need to consider for their specific robot and tasks}.

Additionally, as seen in Fig.~\ref{fig:attempts}, the accuracy of our speech interface for certain feeding tasks was subpar, especially on first attempts by participants. Consistency of the interface is an important consideration for LLM speech interfaces as presented in our design guidelines, Fig.~\ref{fig:design_guidelines}, and explained in Section~\ref{consistency}. Since the testing of our interface, which used OpenAI's GPT-3.5 Turbo model, there have been several advancements in LLMs and their performance. We urge other researchers to test out various closed and open-source LLMs and benchmark their performance as speech interfaces before selecting one. Fine-tuning of the LLM using commands gathered from study participants could also help address subpar accuracy~\cite{kim2024understanding, vemprala2023grid}.

\section{Conclusion}
In this work, we presented an iteratively constructed framework that emphasizes both prompt/system engineering and human-centric considerations for integrating LLMs as speech interfaces for assistive robots. We developed a speech interface using OpenAI's GPT-3.5 Turbo for the Obi feeding robot and used the system to iterate on the proposed framework. We conducted a final human study with our LLM speech interface for the Obi feeding robot at an independent living facility with 11 older adults and used both qualitative and quantitative data from this study to develop design guidelines. We urge researchers, engineers, and designers to refer to our framework and design guidelines in order to develop robust and human-centered LLM-based speech interfaces for assistive robots.  

\begin{acks}
\change{This research was supported by the National Science Foundation Graduate Research Fellowship Program under Grant No. DGE1745016 and DGE2140739. We thank Jon Dekar and Mike Miedlar from DESĪN LLC for providing support with the Obi robot. We also thank the staff and residents of Baptist Providence Point independent living facility for their support of our human study.}
\end{acks}

\bibliographystyle{ACM-Reference-Format}
\bibliography{references}

\clearpage
\appendix

\setcounter{table}{0}
\renewcommand{\thetable}{A\arabic{table}}
\renewcommand{\thefigure}{A\arabic{figure}}
\setcounter{figure}{0}

\section{Framework Iteration Details}
\subsection{Version 1 Pilot Testing}
\label{appendix:v1pilottesting}

For our version 1 pilot testing, four lab members, separate from the immediate research team, were presented with short videos depicting the robot performing four tasks which were (1) feeding a single scoop from one bowl, (2) feeding multiple scoops from one bowl in succession, (3) feeding scoops from multiple, different bowls in succession, and (4) scraping down the sides of a bowl to move food to the center of the bowl. They were then asked to each provide three different ways of phrasing commands to the robot to complete each task, intending to gather a broader set of phrasings that was not influenced by the researcher's vocabulary; this resulted in a total of 48 phrasings. Lastly, we gauged all individuals on improvements and features they would like to see implemented in the system. 

We tested the accuracy of the LLM using the 48 collected command phrasings. We recorded which phrasings did not elicit the intended code from the LLM, and we subsequently analyzed these mistakes to determine if there were common patterns. \change{The intended code was defined as the code that completed the task without any unnecessary additions from the LLM}. Next, we made minor modifications to the prompt, tackling one mistake at a time. Subsequently, we tested the 48 commands with the new prompt and compared the LLM's new output to the output from the previous prompt to assess whether our modification rectified the issue and whether it introduced any unforeseen side effects. We repeated this process of testing, analysis, and modification multiple times until all of the mistakes were addressed. 

After this process, instead of using recorded videos of robot tasks, three members of the research team directly interacted with the Obi robot, allowing us to confirm both the accuracy of the LLM's output and additionally assess the overall dining experience. We filled three of Obi's bowls with yogurt, Cheerios cereal, and blueberries and attempted to use the robot to complete each of the four tasks mentioned above. We followed a similar analysis and prompt modification process as discussed above, but with an additional focus on identifying ways in which the prompt could be altered to enhance user comfort during the assisted feeding. Finally, we similarly solicited feedback from the three members of the research team regarding enhancements for the system. 

\subsection{Version 2 Framework Iteration}
\label{appendix:v2frameworkiter}

Based on our pilot testing with lab members and the research team, we identified that the Code Specifications in our prompt were not sufficiently detailed, as GPT would frequently attempt to import modules that had already been imported or did not exist, attempt to call functions that did not exist, define unnecessary new functions to carry out the user's request, or simply not generate code. Thus, we modified the description of Code Specifications in the framework to express the importance of explicitly outlining how GPT should write code. 

We noticed that GPT would sometimes call robot functions \change{from the Robot API} incorrectly, primarily in the following three ways. First, if a user requested a scoop of food, GPT would frequently scoop from a bowl without bringing the food up to the user's mouth. Additionally, if a user said ``scrape down the yogurt bowl and then feed me some'', GPT would frequently scrape down the sides of the bowl and then immediately move to the user's mouth, without scooping to actually pick up a bite. Lastly, if a user asked to be fed multiple times in succession, GPT would frequently not pause long enough at the user's mouth for the user to comfortably take the bite, instead moving immediately from the user's mouth to the next bowl to scoop from. For these reasons, we expanded the Function Applications component of the framework to emphasize the need to not only describe what each of the functions do, but to describe how the functions relate to each other and to provide examples of how the functions should be combined. To further alleviate these issues, we modified the Robot Functions component, to suggest the merging of functions that are always called in sequence. These solutions generalize to other assistive robots: for example, a mobile manipulator in the home may need to do multiple actions in sequence, such as grabbing a cup and filling it with water before bringing it to the user. For this case, if GPT frequently skipped the filling step, specifying the logical order of functions in the prompt or even merging the ``grabbing cup'' and ``filling water'' functions would lead to better performance. 

%Furthermore, we iterated on the description of Robot Functions to emphasize the importance of defining simple functions and merging functions that are always called in sequence.

\subsection{Version 2 Implementation}
\label{appendix:v2implementation}
Using the revised Code Specifications component as a guide, we adjusted our prompt to emphasize the necessity of producing responses exclusively in Python code, refraining from importing extra modules, and utilizing solely the provided functions. Additionally, we specify that the LLM should not create new functions or use hypothetical ones.

To adhere to the updated Function Applications component, we added more information to the prompt about how the robot functions should be combined. Specifically, we placed additional emphasis on Obi moving to the user's mouth if the user asked to be fed or have a bite. Additionally, we specifically instructed GPT to pause between bites if users asked to be fed multiple times in succession to give users the chance to eat each bite; without this pause, GPT would immediately move from the user's mouth to the next scoop and therefore combine the functions incorrectly as the user would not be able to be fed properly. We set the default delay to be 4 seconds and users could customize this delay if desired. 

To address the altered Robot Functions component, the function to scrape down the bowl was modified to always scoop from the same bowl immediately after to enhance GPT's accuracy. This change was incorporated as we realized whenever a user requested that Obi scrape down a bowl, it was with the intention of scooping food from that bowl immediately after. Previously, GPT often failed to comprehend this sequence, leading to instances where it would incorrectly scrape down the bowl's sides and then move directly to the user's mouth without scooping, despite the user's intention for Obi to scoop as well.

Lastly, to incorporate the new Robot Variables component, we defined three new global variables that affect the way in which Obi carries out its movements. These include \verb|obi.speed|, which sets the speed at which Obi moves and ranges from 0 degrees per second (slowest) to 80 (fastest); \verb|obi.accel|, which determines the rate at which Obi accelerates and ranges from 0 degrees per second squared (slowest) to 250 (fastest); and \verb|obi.deep_scoop|, which dictates the size of the scoops Obi takes with bigger scoops if true and smaller scoops if false. The speed ranges and scoop depths were chosen \change{through experimental testing within the research team} with the modified Safety component in mind; specifically, in terms of speed, we found that moving the robot at full speed (120 degrees per second) presents potential risks, such as food spillage. \change{To further adhere with the updated Safety component, we introduce an LLM output processing step to clip speeds to ensure they are within the set bounds}. Lastly, we added the names of each of these robot variables to the prompt, along with descriptions of what they affect, their ranges, and their default values.

\subsection{Version 3 Framework Iteration}
\label{appendix:v3frameworkiter}

First, we identified that adjustments were needed for the Robot Variables component due to GPT's inconsistent modification of continuous variables spanning wide ranges. As an example, for the Obi system, when the LLM was provided a robot actuator speed range of 0 to 80 degrees per second and commanded to ``move a little faster'', it would sometimes increase the speed by 5 degrees per second and sometimes by 20 degrees per second. We found that users would often not notice a change in speed for small increments and were confused if the speed increment was too large. This process unveiled that GPT lacked physical understanding of robot variables, but could be grounded by using small, discrete ranges for variables. For example, in the case of actuator speed for Obi, we could ground GPT's understanding of what was ``fast'' or ``slow'' in a 0-5 discrete scale leading to speed adjustments that would be more consistent and in line with user expectations. Thus, we modified the Robot Variables component accordingly to suggest the grounding of physical variables in the prompt into small, discrete ranges with default values.

\subsection{Version 3 Implementation}
\label{appendix:v3implementation}

As discussed in the main text, for the Robot Variables component, we changed the speed and acceleration ranges to discrete ranges. The outputted values from GPT were linearly scaled to the original ranges before execution of the code on the robot. 

In accordance with the new Feedback component, we made significant changes in the way users interacted with the system. Instead of holding down a button when they wanted to command the robot, users would now say a wakeup phrase, ``Hey Obi,'' to indicate that they were about to begin a command. The wakeup was implemented using the Porcupine Python API from PicoVoice~\cite{porcupine}. Additionally, audio cues played by the system were added: the system would play a beep when it registered the user saying the wakeup phrase. It would additionally say ``Got it, processing,'' after the user finished speaking to indicate it had heard the user's request and would say ``Scooping now'' or ``Scraping now'' whenever it was about to execute the respective action so that users would not be surprised by the robot’s movement. Lastly, the robot would announce ``Ready for another command'' once it had finished executing code for the user's request. 

Likewise, to satisfy the Instructional Materials component of the framework, we developed more detailed and standardized resources for learning how to use the system. We recorded an instructional video, included as Supporting File 1, explaining the purpose of the robot, how to talk to it, what to expect when using it, and what capabilities it has. The video also showed a simple demonstration of using the robot so users could see what eating with the robot looked like before trying it themselves. Additionally, we made a reference sheet, included as Supporting File 2, to summarize the capabilities of the robot. The intended purpose of this sheet was for users to have it on hand while using the robot in case they were unsure what the robot could do or how to communicate with the system. These instructional materials were intended to ease users into the new technology, with the expectation that these materials were temporary and wouldn't be needed as users became more comfortable with the system and its functions.

\section{Questionnaires}
\label{appendix:questionnaires}
Participant responses to the below questionnaires were examined by researchers to identify any trends or insights. Some responses were used as quotes in the main text.
\subsection{Demographics and Pre-Study Questions} 
These were asked prior to introducing Obi and the speech interface or any of its capabilities to the participant.
\begin{enumerate}
    \item What is your gender?
    \item What is your age?
    \item What is your ethnicity?
    \item How much experience do you have interacting with robots (where 1 is no experience and 7 is expert)?
    \item How do you feel about using robots to help with everyday tasks (where 1 is very negatively and 7 is very positively)?
\end{enumerate}
\subsection{Predefined Task Questions}
These questions were asked after each predefined task and the initial practice task. \\
The participants were asked to answer the following question with yes or no
\begin{enumerate}
    \item Did the robot adequately complete the intended task?
\end{enumerate}
The participants were asked to respond to the following Likert items on a seven-point scale where 1 is strongly disagree, 2 is disagree, 3 is somewhat disagree, 4 is neither agree nor disagree, 5 is somewhat agree, 6 is agree and 7 is strongly agree.
\begin{enumerate}
    \item I found it easy and straightforward to accomplish the specified task.
    \item I didn't have to put in any mental effort to complete the task.
    \item I did not feel frustrated while completing the task.
\end{enumerate}
\subsection{System Usability Scale}
Participants were asked to respond to the following System Usability Scale questions on a 5-point scale where 1 is strongly disagree and 5 is strongly agree at the end of the study.
\begin{enumerate}
    \item I think that I would like to use this system frequently.
    \item I found the system unnecessarily complex.
    \item I thought the system was easy to use.
    \item I think that I would need the support of a technical person to be able to use this system.
    \item I found the various functions in this system were well integrated.
    \item I thought there was too much inconsistency in this system.
    \item I would imagine that most people would learn to use this system very quickly.
    \item I found the system very cumbersome to use.
    \item I felt very confident using the system.
    \item I needed to learn a lot of things before I could get going with this system.    
\end{enumerate}
\subsection{Likert Items}
\label{appendix:likert}
Participants were asked to respond to the following Likert items on a 7-point scale where where 1 is strongly disagree, 2 is disagree, 3 is somewhat disagree, 4 is neither agree nor disagree, 5 is somewhat agree, 6 is agree and 7 is strongly agree at the end of the study.
\begin{enumerate}
    \item It was easy to learn how to speak to the robot.
    \item It was easy to control the robot by speaking to it.
    \item I was able to control the robot through speech in a reasonable amount of time.
    \item The robot was enjoyable to use.
    \item I felt in control when using the robot.
    \item If I was unable to eat independently, I would use this robot and speech interface to feed myself.
\end{enumerate}
\subsection{NASA Task Load Index (TLX)}
Participants were asked to respond to the following NASA TLX questions on a 100-point scale where 0 is low workload and 100 is high workload at the end of the study.
\begin{enumerate}
    \item How mentally demanding was the task?
    \item How physically demanding was the task?
    \item How hurried or rushed was the pace of the task?
    \item How successful were you in accomplishing what you were asked to do? (here 0 indicates perfect and 100 indicates failure)
    \item How hard did you have to work to accomplish your level of performance?
    \item How insecure, discouraged, irritated, stressed, and annoyed were you?
\end{enumerate}
\subsection{Open-ended Questions}
\label{appendix:open-ended}
These questions were asked to participants at the end of the study
\begin{enumerate}
    \item What did you like about talking to the robot? What did you dislike?
    \item Did you feel limited in your ability to command the robot through speech? If so, how?
    \item Did the robot make mistakes or misunderstand you? If so, how?
    \item What features/capabilities would you like to see from the robot that it doesn’t have already?
\end{enumerate}

\section{Participant Commands}
\label{appendix:participant_commands}
For the predefined tasks, we include all the participant commands generated by the Whisper speech to text API. If applicable, we nest the second and third commands for each task and denote if the third command was successful (S) or not successful (U). 

As a reminder, success was reported by the participant and this led to some discrepancies. Specifically, there are some noticeable issues in some of the commands due to the following reasons:
\begin{itemize}
\item Whisper speech to text translation errors
\item Participants not speaking clearly 
\item Participant not fully understanding task instructions
\item Missing the first few words of the command as the participant spoke before the "beep" after the "Hey Obi" wakeup word was detected
\item Even if the correct code was generated and executed, sometimes the participants thought the robot didn't change its behavior. The vice versa sometimes occurred where wrong code was generated but the participant thought the robot changed its behavior. This was especially the case for Task 1 and 2 related to speed and scoop size commands.
\end{itemize}

\begin{figure*}[hbt!]
  \centering
  \includegraphics[width = \textwidth]{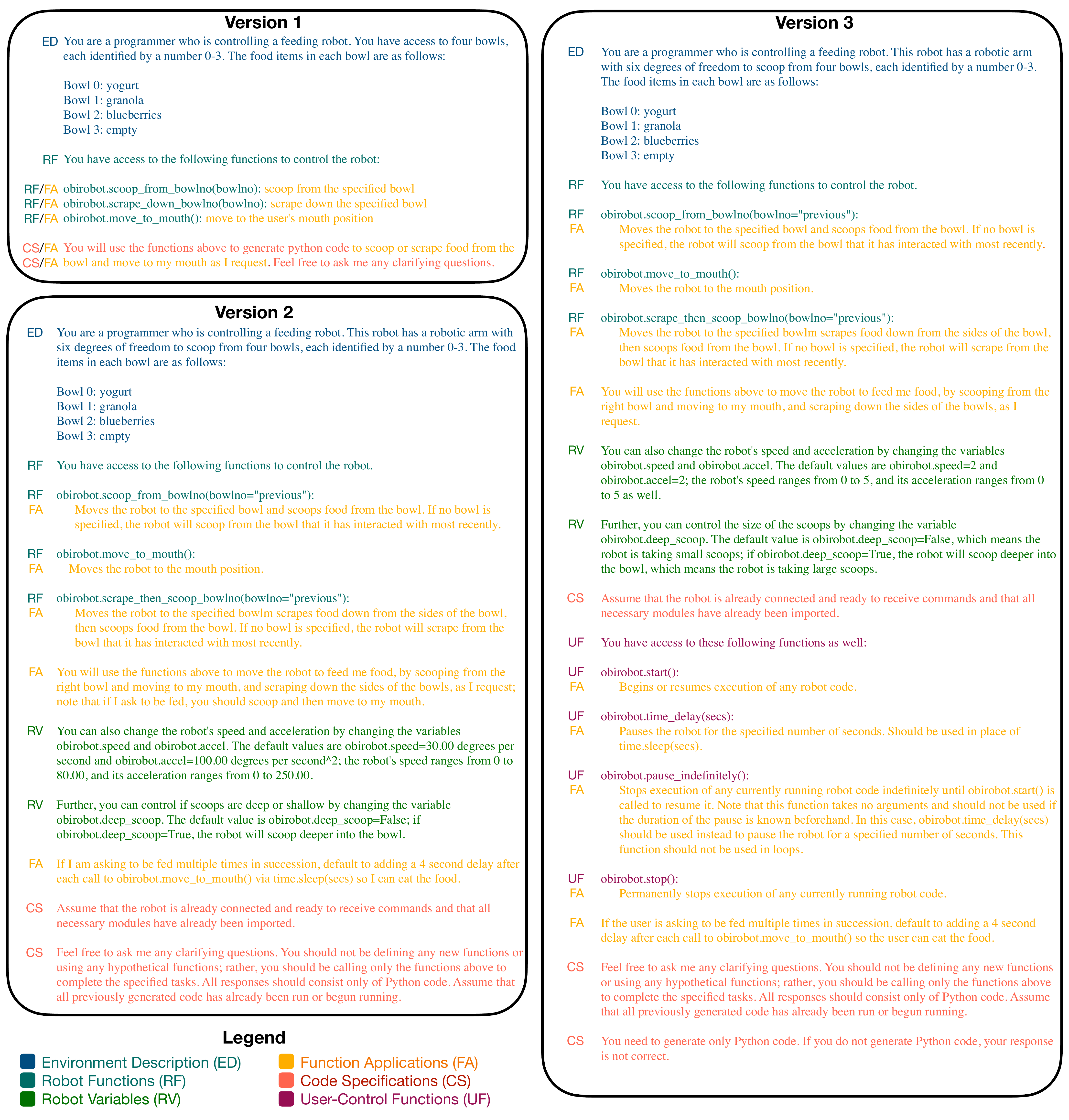}
  \caption{Prompt Iteration. The three versions of our prompt are shown with the label and text color indicating the framework component that is being satisfied.}
  \Description{Prompt Iteration. The three versions of our prompt are shown with the label and text color indicating the framework component that is being satisfied.}
  \label{fig:prompt_iteration}
\end{figure*}

\change{
\subsection{Practice Task}
\begin{itemize}
    \item Phoenix or yogurt.
    \item Phoebe One Scoop of Yogurt.
    \item Feed me some yogurt.
    \item Please give me a scoop of yogurt.
    \item Go ahead and try that again. Yeah, good. OK. Hey, Opie. Give me some blueberries.
    \item May I have a bite of pudding?
    \begin{itemize}
        \item May have I have pudding, please.
    \end{itemize}
    \item I'd like a scoop of blueberries.
    \begin{itemize}
        \item Give me a scoop of yogurt.
    \end{itemize}
    \item blueberries for breakfast please.
    \item Feed me yogurt.
    \item Feed me a scoop of blueberries.
    \item Cheerias.
    \begin{itemize}
        \item Feed me a scoop of Cheerios, please.
        \item Feed me another scoop of Cheerios. (S)
    \end{itemize}
\end{itemize}

\subsection{Task 1}
\begin{itemize}
    \item Feed me a scoop of granola.
    \item and all them work quickly.
    \begin{itemize}
        \item feed me a scoop of granola.
    \end{itemize}
    \item Feed me some granola faster.
    \item Scoop of Granola.
    \begin{itemize}
        \item A scoop of granola.
        \item Please feed me granola quickly. (S)
    \end{itemize}
    \item Give me um, Cheerios quickly.
    \begin{itemize}
        \item Cheerios quickly.
        \item No, no, you're all good. Go ahead. Hey, Obie. Give me Cheerios quickly. (U)
    \end{itemize}
    \item Feed me the yogurt a little faster.
    \item Give me a scoop of granola.
    \item Quickly feed me some granola.
    \item granola quickly.
    \begin{itemize}
        \item faster.
        \item Feed me granola. (S)
    \end{itemize}
    \item Quickly feed me a scoop of pretzels.
    \item We'll use.
    \begin{itemize}
        \item Will you feed me some granola and do it faster?
        \item Will you feed me some granola and do it faster? (S)
    \end{itemize}
\end{itemize}

\subsection{Task 2}
\begin{itemize}
    \item Feed me a bigger scoop of blueberries.
    \item Feed me a bigger scoop of blueberries.
    \item Feed me some blueberries slower.
    \item Scoop of blueberries.
    \begin{itemize}
        \item A large scoop of blueberries.
    \end{itemize}
    \item Give me a bigger scoop of... Hey, Obi, give me a bigger scoop of Cheerios.
    \item Give me a bigger scoop of apple sauce.
    \begin{itemize}
        \item Give me another scoop of apple sauce.
        \item Please give me another scoop of apple sauce. (S)
    \end{itemize}
    \item blueberries.
    \item Quickly me, I have some blueberries.
    \item Feed me a big scoop of blueberries.
    \item Feed me a large scoop of blueberries.
    \item Feed me a scoop of Cheerios and make it quicker, larger. I'm sorry. I confused Obi.
\end{itemize}

\subsection{Task 3}
\begin{itemize}
    \item the yogurt to the side.
    \begin{itemize}
        \item Feed me the yogurt.
    \end{itemize}
    \item Move the yogurt back to the center of the bowl and feed me one scoop.
    \begin{itemize}
        \item down the sides of the yogurt bowl and feed me wet scoop.
        \item down the sides of the yogurt bowl. Feed me one scoop. (U)
    \end{itemize}
    \item Scrape down the yogurt dish and give me a spoon.
    \item scrape down the sides of the yogurt and give me a scoop.
    \item down the side of the apple sauce.
    \begin{itemize}
        \item the OB scraped down the sides of the applesauce.
        \item down the sides of the applesauce four times. (U)
    \end{itemize}
    \item Please scrape down the side of the bow of the apple sauce.
    \item the side of the yogurt bowl and give me a spoonful.
    \item scrape down the sides of the yogurt bowl.
    \item the sides of the bowl, yogurt bowl.
    \item down the sides of the yogurt bowl and then feed me a scoop of yogurt.
    \item Feed me some yogurt and scrape down the sides of the bowl.
\end{itemize}

\subsection{Task 4}
\begin{itemize}
    \item Feed me a scoop of canola.
    \begin{itemize}
        \item Feed me a scoop of granola. Three scoops of granola.
    \end{itemize}
    \item Feed me three small scoops of yogurt. Sorry, three small scoops of granola.
    \begin{itemize}
        \item Feed me three small scoops of granola.
    \end{itemize}
    \item Give me three scoops of granola.
    \item three scoops of yogurt.
    \begin{itemize}
        \item The rescoops of blueberries.
        \item three scoops of blueberries. (U)
    \end{itemize}
    \item Cheerios three times quickly.
    \begin{itemize}
        \item areas three times quickly.
        \item I want Cheerios three times very quickly. (U)
    \end{itemize}
    \item Give me please three scoops of applesauce.
    \item Renola.
    \begin{itemize}
        \item I'd like three scoops of yogurt.
    \end{itemize}
    \item Three scoops of blueberries, please.
    \begin{itemize}
        \item Three scoops of blueberries.
        \item I would like to have three scoops of blueberries in a row. (S)
    \end{itemize}
    \item Feed me three scoops of blueberries.
    \item Feed me three small scoops of blueberries.
    \item Feed me three scoops of Cheerios, please.
\end{itemize}

\subsection{Task 5}
\begin{itemize}
    \item Feed me a scoop of blueberries, followed by a scoop of yogurt.
    \item blueberries and then a scoop of yogurt.
    \begin{itemize}
        \item Feed me a scoop of blueberries and a scoop of yogurt.
    \end{itemize}
    \item May I have a scoop of berries and a scoop of granola?
    \item One scoop of blueberries and one scoop of granola.
    \begin{itemize}
        \item scoop of blueberries and a scoop of yogurt.
    \end{itemize}
    \item Give me one scoop of blueberries and one scoop of jureos.
    \begin{itemize}
        \item I want you to fill the blueberry, fill a spoon with blueberries and a spoon with Cheerios and make it full.
        \item I want blueberries and Cheerios quickly and make it full. (S)
    \end{itemize}
    \item Give me a scoop of apple sauce and then a scoop of pudding.
    \item I'd like a scoop of granola and a scoop of blueberries.
    \item I'd like some granola and some yogurt please.
    \begin{itemize}
        \item I'd like some blueberries and some yogurt please.
    \end{itemize}
    \item granola, then yogurt.
    \item and then feed me a tiny scoop of yogurt.
    \begin{itemize}
        \item me a scoop of pretzels and then feed me a tiny scoop of yogurt.
    \end{itemize}
    \item Feed me a scoop of granola and a scoop of yogurt.
\end{itemize}
}

\begin{table}[ht]
    \centering
    \begin{tabular}{|c|c|c|}
        \hline
        \textbf{Participant} & \textbf{Time between Bites (s)} & \textbf{Number of Bites}\\
        \hline
        1 & 38$\pm$7 & 12 \\
        2 & 50$\pm$15 & 8 \\
        3 & 42$\pm$14 & 8\\
        4 & 32$\pm$6 & 14\\
        5 & 55$\pm$16 & 7\\
        6 & 42$\pm$14 & 11\\
        7 & 37$\pm$8 & 11\\
        8 & 40$\pm$15 & 11\\
        9 & 71$\pm$25 & 7\\
        10 & 48$\pm$17 & 10\\
        11 & 44$\pm$11 & 10\\
        \hline
        \textbf{Mean$\pm$SD} & 45$\pm$10 & 10$\pm$2 \\
        \hline
    \end{tabular}
    \caption{Time between Bites and Number of Bites per Participant}
    \label{tab:avg_bite_timing}
\end{table}

\section{Bite Timing Metrics}
\label{appendix:bite_timing_metrics}
\change{From the open feeding session from our study with older adults, we present the average bite timing for each participant in Table~\ref{tab:avg_bite_timing}. Researchers in robot-assisted feeding should interpret these values cautiously, as the bite timing could have been prolonged due to the participants needing to issue the speech commands and due to the speed of the robot. We additionally noticed that participants who gave multi-step instructions as commands had shorter bite timings. }

\end{document}